\documentclass[11pt]{article}

\usepackage[margin=1in]{geometry}
\usepackage[american]{babel}
\usepackage[T1]{fontenc}
\usepackage[utf8]{inputenc}
\usepackage{lmodern}
\usepackage{microtype}
\usepackage{csquotes}
\usepackage{graphicx}
\usepackage{xurl}
\usepackage{amsmath}
\usepackage{array}
\usepackage{tabularx}
\usepackage[table]{xcolor}
\usepackage{booktabs}
\usepackage{pdflscape}
\usepackage{placeins}
\usepackage[font=small,labelfont=bf]{caption}
\usepackage[hidelinks]{hyperref}
\usepackage[style=apa,sortcites=true,sorting=nyt,backend=biber,natbib=true]{biblatex}
\DeclareLanguageMapping{american}{american-apa}
\newcolumntype{Y}{>{\raggedright\arraybackslash}X}
\title{Heterogeneous Neural Predictivity from Language Models During Naturalistic Comprehension}
\author{Xiao Jia\\
School of Artificial Intelligence\\
The Chinese University of Hong Kong, Shenzhen\\
\texttt{xiaojia@link.cuhk.edu.cn}}
\date{June 25, 2026}

\begin{document}
\maketitle

\begin{abstract}
Language-model representations provide structured, high-dimensional annotations of naturalistic language stimuli and can serve as informative neural predictors during comprehension. We analyzed locked derived data from Brain Treebank, MEG-MASC, and Podcast ECoG with eight frozen language models, blocked encoding models, and matched temporal, nuisance, and representation-capacity controls. Positive held-out prediction and gains over low-level baselines were widespread in source-level summaries. Across Brain Treebank and Podcast ECoG, 67 of 432 evaluable rows met a controlled predictive-only criterion, and model-side feature ablations changed prediction scores in most evaluable source rows. Brain-derived, timing-linked, acoustic, and implanted-signal controls confirmed component-level sensitivity of the analysis pipeline. These findings show that language-model-derived quantities can annotate neural activity during natural speech and text comprehension. Participant-level matched-control advantages were localized rather than uniform, response-profile and feature-specificity contrasts bounded representational or computational interpretations, and complete co-indexed integrated interpretation will require future jointly indexed coverage. Together, the analyses identify language-model features as useful neural predictors and separate predictive usefulness from claims about shared neural organization or language-processing computations.
\end{abstract}


\bigskip
\section{Introduction}

Language-model representations have become effective quantitative probes of naturalistic language comprehension. Contextual embeddings, surprisal estimates, and layer-specific features can predict neural responses measured with fMRI, MEG, EEG, and intracranial electrophysiology, providing a tractable way to relate unfolding linguistic context to neural activity \citep{huth2016,schrimpf2021,caucheteux2022,goldstein2022,toneva2019,tuckute2024,pereira2018,wehbe2014,hosseini2024nol,jain2024insilico}. This predictive success is scientifically useful because language-model features provide structured, high-dimensional representations of the stimulus that can account for variation in neural responses. The next question is what kind of neural and cognitive information this predictivity supports.

Positive neural prediction can support several levels of inference. A feature may predict held-out neural responses because it tracks stimulus properties relevant to the measurement. Additional evidence can show whether the trained representation exceeds matched temporal, nuisance, and capacity controls, whether it reproduces organization across sampled neural units, or whether it depends selectively on a candidate language-related quantity. The present study treats these outcomes as separable empirical claims.

Naturalistic stimuli make this separation important. Word onset, word rate, acoustic envelope, sentence position, discourse progression, lexical frequency, token predictability, and local transition statistics are intercorrelated. Neural measurements also contain temporal autocorrelation, and modern language-model features are high-dimensional, layer-indexed, and context-length-dependent \citep{vaswani2017,devlin2019,brown2020,qwen25,qwen3}. Positive neural prediction may therefore combine language-related information with temporal, lexical, acoustic, and representation-capacity contributions.

Recent language-neuroscience work motivates a more explicit link between computational measures and cognitive interpretation. Theoretical conclusions depend on the chain connecting constructs, tasks, measurements, analyses, and auxiliary assumptions \citep{vanderburght2023brickyard}. Syntax--semantics, surprisal, and narrative-comprehension studies show the value of linking computational measures to specified neural response patterns and interpretive targets \citep{shain2024syntaxsemantics,weissbart2020surprisal,thye2024narrative}. Hadidi and colleagues further show that shuffled training and testing partitions, activation-extraction choices, positional signals, and word-rate controls can strongly affect brain--language-model predictivity \citep{hadidi2026spurious}. These results motivate a narrower question: how much of the observed signal supports predictive usefulness, model-specific advantage, shared neural organization, or candidate-computation interpretation.

Here we characterize heterogeneous neural predictivity from language-model features across three naturalistic language datasets. We quantify positive neural predictions, gains over nuisance baselines, controlled predictive-only rows, participant-level consistency, and sensitivity to model-side ablation. We then compare these outputs with matched controls, response-profile tests, feature-specificity diagnostics, and reliability-bounded summaries to determine which interpretations are supported by the available derived data.

\section{Materials and Methods}

\subsection{Participants and data sources}

This secondary analysis used preprocessed and derived data from previously released sources; raw neural recordings were governed by the original data providers and were not redistributed. Three datasets were treated as the primary naturalistic language sources because the available derived data contained neural time series or neural targets, word-level event grids, language-model features, nuisance variables, matched controls, and reliability or coverage metadata \citep{braintreebank2024,zada2025podcast,gwilliams2023megmasc}. Brain Treebank contributed 10 participants, 26 subject-run units, and 248 modality-specific brain units. Podcast ECoG contributed 9 participants, 9 subject-run units, and 235 ECoG-derived brain units. MEG-MASC contributed 11 participants, 84 subject-run units, and 257 MEG-derived target units. These brain units are electrodes, sensors, source targets, time windows, or derived target profiles depending on the source dataset. They are distinct from independent participants.

The sample size was determined by the secondary-data design and by the subset of publicly released or locally accessible derived data that could be matched across neural targets, word-event grids, model features, controls, and reliability metadata. No new participants were recruited, and participant expansion was outside the available derived-data scope. Inference is therefore bounded by the participant and participant-run coverage retained for each contrast. The matched participant-run predictive inference retained 26 Brain Treebank subject-run units from 10 participants, 44 MEG-MASC subject-run units from 11 participants, and 8 Podcast ECoG subject-run units from 8 participants after complete model-control matching. When a participant contributed multiple runs, runs were averaged within participant before participant-cluster bootstrap. These coverage counts define the predictive inference boundary; electrodes, sensors, layers, and table rows define nested analysis dimensions.

The source datasets contain material beyond the subset with complete local feature matching. Brain Treebank records intracranial electrophysiology while neurosurgical participants watched naturalistic movie stimuli, with manually corrected transcripts, word onsets, part-of-speech labels, and dependency parses aligned to the audio track \citep{braintreebank2024}. Podcast ECoG records intracranial responses while participants listened to a natural spoken podcast, with high-gamma preprocessed derivatives and linguistic feature annotations available from the source release \citep{zada2025podcast}. MEG-MASC records English-speaking participants listening to naturalistic MASC stories across repeated MEG sessions that also include word-list and comprehension-question material \citep{gwilliams2023megmasc}. The present manuscript analyzes the participant-run, event-grid, representation, and control rows that could be matched in the derived data.

The source inventory retained Narratives and LPP multilingual as secondary or exploratory sources \citep{nastase2021,li2022lpp}. Learning Brain was treated as validation-only, and Natural Stories was treated as stimulus--language-model-only or diagnostic because the available derived data lacked the neural-side coverage required for the present contrasts \citep{futrell2021naturalstories,learningbrain2025}. Participant demographics, original exclusion criteria, ethics approvals, and consent procedures are governed by the source publications and repositories. The present manuscript reports the derived units available to this analysis and preserves de-identification and raw-data access boundaries.

\subsection{Naturalistic stimuli and neural measurements}

The three primary datasets sample naturalistic comprehension with different measurement modalities. Brain Treebank provides intracranial recordings during naturalistic audiovisual or language stimuli. Podcast ECoG provides intracranial recordings during podcast comprehension. MEG-MASC provides MEG responses to natural speech. The local analysis inherited each source's preprocessing, artifact rejection, and target definition from the released or derived artifacts. Word onset and event grids were used to align language-model features and nuisance variables to the neural targets.

The primary inferential unit was the participant or participant-run whenever the matched data allowed that aggregation. Electrodes, sensors, target windows, layers, models, metrics, and candidate quantities were treated as nested or crossed analysis dimensions. Row counts describe coverage of model--dataset--layer--metric combinations; independent predictive evidence comes from participant-aware summaries. A modality, region, and time-window summary was generated from the predictive outputs. In the matched derived data, predictive rows retain modality labels and a broad target-coverage label. Modality and window summaries are therefore available, whereas within-region comparisons require finer retained target strata. Supplementary Table 30 lists the dataset-specific neural target type, brain-unit count descriptor, temporal window, selection rule, final unit, and what was averaged before the manuscript-facing contrasts. The broad target label denotes the retained target coverage in the derived data. Predictive intervals are reported over 10 Brain Treebank participants and 26 subject-run units, 11 MEG-MASC participants and 44 subject-run units, and 8 Podcast ECoG participants and 8 subject-run units. Response-profile and feature-ablation summaries retain their contrast-specific target-profile or diagnostic scopes and are reported as bounded summaries without formal participant-level equivalence tests.

\subsection{Language-model representations and candidate quantities}

The analysis used fixed language-model representation files from the analysis package. The analyzed model inventory was bounded to eight validated feature sets: DistilGPT-2, GPT-2, GPT-2 Medium, Pythia-160M, Pythia-410M, Qwen2.5-0.5B-Instruct, Qwen2.5-1.5B-Instruct, and Qwen3-1.7B. Larger locally indexed checkpoints without matched analysis rows, including Qwen2.5-7B-Instruct and Qwen3-4B-Instruct-2507, were outside the analyzed model set.

Candidate language-related quantities were operationalized from the fixed representation files and word-event tables. Word surprisal was computed in natural-log units as the summed subword surprisal for each word, \(-\sum_{i \in w}\log p(t_i \mid t_{<i})\), using model logits and token-to-word maps. The event-level implementation used model-derived surprisal when available and otherwise used the fixed unigram-surprisal proxy; proxy rows are treated as source-boundary diagnostics. Proxy-derived quantities are excluded from model-specific computational correspondence claims. Semantic transition and context update were read from the available word-feature annotations. Dependency integration, syntactic boundaries, and discourse boundaries were inherited from the event grid as scalar feature streams or event-indicator streams. These operational variables annotate contextual predictability, semantic change, syntactic or dependency structure, boundary structure, and context updating. Biological-mechanism claims require evidence beyond these operational variables alone.

\subsection{Temporal alignment, encoding models, and matched controls}

Where held-out scores were recomputed from matchable event, feature, and neural grids, neural prediction used ridge-regularized linear encoding within blocked cross-validation \citep{hoerl1970,hastie2009,pedregosa2011}. Lag structure, ridge penalty, PCA dimensionality, residualization, projection-removal steps, and ablation transformations were selected or fit only inside training data, following leakage-prevention cautions from predictive modeling and neuroimaging \citep{stone1974,varma2006,varoquaux2017,yarkoni2017,kriegeskorte2009}. The blocked design was used to reduce temporal leakage from autocorrelated naturalistic stimuli. Supplementary Table 29 summarizes the manuscript-facing implementation settings, including block and fold policy, ridge alpha grid, lag and time-window specification, PCA dimensionality cap, hidden-state token pooling, subword-to-word aggregation, score aggregation unit, and proxy policy.

Matched controls tested whether a model advantage was specific to the real language-model representation. The control families included nuisance features, random matched-dimensionality features, autocorrelation-matched random features, circular shifts, sentence-reset features, reversed-context features, layer-label permutation, token-order shuffle, and within-story block shuffle. For each model--dataset--layer--metric summary row, the matched-control sensitivity contrast compared the real-model score with the most competitive available matched control for the same row. This comparison asks whether a result remains positive after the closest available alternative in the matched derived data. Family-specific contrasts, the frequency with which each family became the most competitive control, and single-control-family removal analyses are retained in the Supplement and detailed CSV files to show the role of individual control families.

\subsection{Claim hierarchy and statistical summaries}

The analysis separates information-bearing predictivity from model-specific predictive advantage. Information-bearing predictivity refers to positive held-out prediction or improvement over a low-level nuisance baseline. It establishes that a representation contains information useful for neural prediction, with model specificity evaluated at the next evidence level. Model-specific predictive advantage requires a positive real-minus-matched-control contrast under the configured temporal, capacity, and contextual controls. Cross-neural-unit response-profile correspondence is a representational-organization contrast asking whether model-to-brain profiles reproduce brain-derived profile organization over the same sampled units. Candidate-computation ablation is a computation-specificity contrast asking whether model-side ablation of a candidate language-related quantity selectively changes held-out neural prediction, conditional on predictive evidence. Reliability-bounded response-profile magnitude asks whether a surviving response-profile effect is large relative to the reliable brain-derived profile signal. Integrated summary rows combine the predictive, response-profile, feature-ablation, reliability-bounded response-profile magnitude, matched-control, and replication criteria.

\begin{table}[!htbp]
\begin{center}
\caption{Claim-to-evidence mapping for language-model neural predictivity.}
\label{tab:claim_evidence}
\begingroup
\scriptsize
\setlength{\tabcolsep}{3pt}
\renewcommand{\arraystretch}{1.08}
\begin{tabularx}{\textwidth}{p{0.18\textwidth}Yp{0.25\textwidth}Y}
\hline
Evidence level & Operational criterion & Current result & Permitted interpretation \\
\hline
\rowcolor[gray]{0.96}
Information-bearing & Positive held-out prediction or gain over a low-level nuisance baseline. & Positive source-level scores and nuisance gains were widespread. & Language-model features show heterogeneous neural predictivity at source and configuration levels. \\
Model-specific & Real-model prediction exceeds nuisance and matched temporal, order, and capacity controls under the configured predictive rule. & 67 of 432 evaluable summary rows met the predictive-only criterion; participant-level mean predictive contrasts were localized rather than uniformly positive. & Local controlled predictive positives are present, with dataset-level averages defining the scope of model-specific advantage. \\
\rowcolor[gray]{0.96}
Representational & Model--brain response profiles exceed shuffled or matched profile controls over sampled neural units. & Raw profile similarities were widespread; dataset-level profile-control contrasts were below the most competitive matched controls. & Shared response-profile organization requires evidence beyond positive raw profile similarity. \\
Reliability-bounded & A surviving response-profile effect is large relative to valid brain--brain profile reliability. & Ceiling-normalized rows were available; reliability-bounded profile contrasts remained below matched controls. & Reliability-bounded profile magnitude is a higher evidence level than positive raw profile similarity. \\
\rowcolor[gray]{0.96}
Computational & Model-side feature ablation shows selective dependence on a candidate quantity, conditional on predictive evidence and feature-specificity diagnostics. & Ablation deltas were often positive, with specificity below the configured criterion. & Candidate quantities are useful diagnostic annotations; computation-specific correspondence requires additional aligned evidence. \\
Integrated & Predictive, representational, computational, reliability-bounded, matched-control, and replication criteria are jointly indexed and supported. & Complete co-indexing across all evidence levels was unavailable in the current derived coverage. & Integrated interpretation requires future datasets or derived files with jointly indexed contrasts. \\
\hline
\end{tabularx}
\endgroup
\end{center}
\vspace{-0.6em}
\noindent\footnotesize Alternative explanations addressed at each level are listed in Supplementary Table 3.
\end{table}
\FloatBarrier

The Supplementary Information documents the thresholded contrast rules used for reproducibility. In the main text, supported, unsupported, and unavailable refer to the specified contrast and available data chain. Integrated summary rows represent the combined claim level and are treated as coverage summaries; participant-level predictive contrasts are reported separately.

\subsection{Response-profile and reliability-bounded response-profile analyses}

Response profiles were constructed to test cross-neural-unit organization beyond target-wise prediction. For each matched dataset, subject, run, stimulus, model, layer, and candidate quantity, the model--brain profile was the ordered vector of held-out readout scores across sampled neural units. Each element of this vector corresponds to one electrode or MEG sensor-group target after event alignment and blocked cross-validated ridge readout. The brain--brain profile used the same unit ordering for brain-derived pattern vectors. Target order was fixed by a stored unit-order hash; rows with mismatched order were marked invalid. Nonfinite profile elements were dropped pairwise for the similarity calculation. Pearson and Spearman profile similarities use their standard centered or rank-centered definitions; cosine similarity uses the finite profile vectors without additional centering. The configured profile readouts are descriptive per-unit readout profiles, distinct from matrix-level CKA or representational-similarity-analysis estimates. In the implemented event-response readout, per-unit Pearson prediction scores are transformed to \(r^2\) for the squared-correlation profile, to \(|r|\) for the absolute-correlation profile, and kept signed for the signed-correlation profile.

For each profile-similarity cell, the real model was compared with matched profile controls generated from the same event grids and target order. The fixed controls include matched-dimensionality random features, autocorrelation-matched random features, circular shifts, context-reset and reversed-context features, layer-label permutations, token-order shuffles, and within-story block shuffles when the corresponding representation files are available. The summary table uses the maximum profile similarity among available controls as the conservative control contrast for that cell; profile controls are deterministic or seeded by the fixed configuration, with no repeated resampling of null draws. The response-profile delta is \(\Delta_{\mathrm{profile}} = s_{\mathrm{real}} - \max_c s_c\), where \(s\) is the configured profile-similarity metric. A response-profile cell can pass when the real profile similarity and control similarity are finite, the target order matches, and \(\Delta_{\mathrm{profile}}>0\). For manuscript-facing summary rows, these profile deltas are then averaged over the sampled unit and subject-run rows in the corresponding dataset--model--layer--candidate-quantity contrast; isolated positive target rows are descriptive within their local target scope. A valid brain ceiling is a requirement for the separate reliability-bounded response-profile magnitude criterion.

Brain reliability ceilings were computed from brain-derived profile vectors. Split-half reliability uses the split-half brain-ceiling values stored with the brain pattern table. Run-to-run, subject-to-subject, and session-to-session reliability, when available, are pairwise Pearson similarities between brain pattern vectors sharing the same dataset, region group, profile-similarity metric, and unit key while differing in the named grouping variable. Method-specific reliability is the mean of available pairwise values, with a 500-sample percentile bootstrap interval. A ceiling is valid only when reliability is finite and at least 0.10. Ceiling-normalized response-profile summaries use \(f_{\mathrm{ceiling}} = s_{\mathrm{real}} / r_{\mathrm{brain}}\) and \(\Delta f_{\mathrm{ceiling}} = \Delta_{\mathrm{profile}} / r_{\mathrm{brain}}\). Negative or missing reliabilities are retained as invalid ceilings. The reliability-bounded response-profile criterion additionally requires \(f_{\mathrm{ceiling}}\ge 0.50\), a positive profile-control delta, and a passing response-profile cell. Predictive uncertainty summaries use 1000 bootstrap samples over participant means. When a participant contributed multiple runs, runs were first averaged within participant and the bootstrap resampled participants at the participant level. When only one participant was retained, participant-cluster inference was marked as unavailable. Feature-ablation double-dissociation tests use 1000 bootstrap and 1000 sign/permutation samples where applicable. False-discovery-rate values use Benjamini--Hochberg correction over the configured dataset-by-candidate-quantity-by-model-by-region family and are interpreted with \(q<0.05\).

\subsection{Model-side feature-ablation analyses}

The feature-ablation analyses recomputed held-out scores after feature zeroing, train-only residualization, train-only projection removal, layer ablation, context reset, or reversed-context transformation. These operations differ in their interpretation. Feature zeroing changes the available feature dimensions directly. Residualization and projection removal remove variance associated with a candidate quantity while preserving train-only fitting. Layer ablation tests dependence on layer-specific representations. Context reset and reversed context test sensitivity to the model's contextual history. For this reason, ablation results are interpreted by operation and candidate quantity before they are summarized across the decision table.

For a candidate quantity \(m\), the feature-ablation delta was \(\Delta_m = s_{\mathrm{real}} - s_{\mathrm{ablated}(m)}\), where \(s\) is the held-out predictive score after the same blocked cross-validation policy. The feature-specificity index was \(\mathrm{FSI}_m = \Delta_m - \overline{\Delta}_{\neg m}\), where \(\overline{\Delta}_{\neg m}\) is the mean ablation delta for nonmatching target quantities within the same dataset, subject-run, region group, time window, model, layer, and ablation method. Double-dissociation tests compared candidate quantities such as surprisal, semantic transition, dependency integration, syntactic boundary, discourse boundary, and context update. They provide diagnostic evidence about selectivity. Neural-system intervention claims are outside the scope of these model-side perturbations. A positive ablation delta is a diagnostic result; integrated interpretation requires predictive matched-control support and response-profile correspondence as separate evidence levels.

\subsection{Positive controls and uncertainty summaries}

Positive controls were stratified by the kind of sensitivity they test. Brain--brain reliability estimates test whether the derived neural data contain repeatable signal. Brain-as-model controls test whether response-profile machinery can recover brain-derived organization. Low-level neural and acoustic checks test component-level sensitivity to timing-linked structure where source media and event grids permit the check. The implanted-signal simulations are integrated engineered controls: a single strong-signal row verifies the implementation under a known signal, and a stochastic graded calibration estimates detection probability across candidate-signal strengths. In the stochastic calibration, synthetic brain responses were generated as \(y=\beta x_{\mathrm{implant}}+\epsilon\) for the implanted latent feature in the affected units; the plotted strength is the coefficient \(\beta\) on a unit-variance latent scale, with synthetic observation noise added separately. The 80\% threshold is therefore an implementation-scale threshold in this engineered parameterization, outside the scale of empirical neural signal-to-noise ratios or empirical predictive and ablation deltas. Podcast brain-as-model summaries are interpreted as auxiliary because they are brain-derived assay checks and are limited to the available subject-to-subject Podcast profile coverage.

Dataset-level uncertainty summaries aggregate over participant-run or derived target units before bootstrapping when those units are available. Supplementary Table 31 reports the aggregation path. The source inventory retains subject, run, stimulus, and brain-unit coverage, and the predictive score table retains subject/session/run/stimulus columns. The participant-run export preserves subject and run identifiers for predictive contrasts where the join is available. The matched predictive inference retained 26 Brain Treebank subject-run units from 10 participants, 44 MEG-MASC subject-run units from 11 participants, and 8 Podcast ECoG subject-run units from 8 participants. Participant-cluster bootstrap resampled participant means after within-participant run averaging. The integrated summary table joins predictive, response-profile, feature-ablation, ceiling, matched-control, and replication summaries by dataset, candidate quantity, model, layer, region group, and time window. This table is treated as a coverage-summary table; population inference is limited to contrasts with retained participant-cluster coverage. Degenerate, low-unit, or low-participant intervals are labeled as coverage limitations. Supplementary Table 18 reports the interval-bound summaries. No smallest effect size of interest was preregistered, so the summaries report observed bounds and excluded-effect limits without formal equivalence tests. Positive upper interval bounds indicate compatibility with positive model-specific effects up to that bound. Resampling, permutation-style diagnostics, and false-discovery-rate correction follow standard bootstrap, nonparametric neuroimaging, and Benjamini-Hochberg logic \citep{efron1994,nichols2002,benjamini1995}. Software versions, model identifiers, random-seed policy, checksums, and provenance records are documented in the Supplementary Information.

\subsection{Human-participant and secondary-data ethics}

This study analyzed de-identified, previously released neural and stimulus-derived data and included no new human-participant recruitment or new human-data collection. As reported in the source publications, Brain Treebank experiments were approved by the Boston Children's Hospital/Harvard Institutional Review Board and were conducted with subjects' informed consent \citep{braintreebank2024}; Podcast ECoG participants provided oral and written informed consent, with study approval from the Institutional Review Boards at New York University Langone Medical Center and Princeton University \citep{zada2025podcast}; and MEG-MASC participants provided written informed consent under approval from the Institutional Review Board ethics committee of New York University Abu Dhabi \citep{gwilliams2023megmasc}. Raw neural datasets and stimulus media were not redistributed and must be obtained from the original repositories or data owners under the terms set by those providers.

\section{Results}

\subsection{Positive predictivity and evidence levels}

\begin{figure}[!htbp]
\begin{center}
\includegraphics[width=0.91\textwidth]{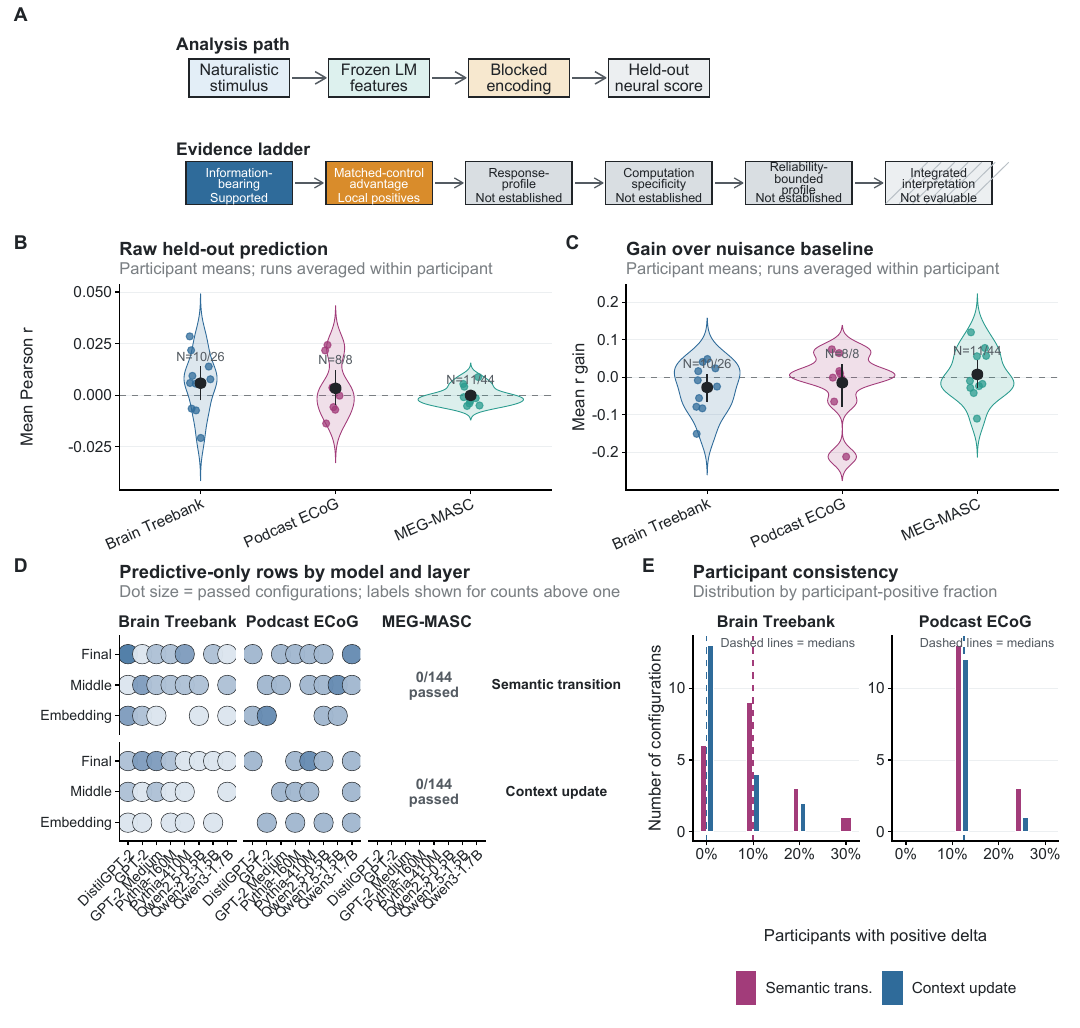}
\caption{Positive information-bearing predictivity and participant-level scope. (A) Analysis path and evidence ladder. (B) Participant-level raw Pearson-\(r\) after within-participant run averaging; intervals are participant-cluster bootstrap summaries. (C) Participant-level gain over the nuisance baseline. (D) Predictive-only criterion rows by dataset, model, layer, and candidate quantity; dot area gives passed configurations, fill gives participant consistency, and labelled empty facets denote zero passed rows among evaluable rows. (E) Participant consistency among predictive-only configurations; dashed lines mark medians.}
\label{fig:positive_information}
\end{center}
\end{figure}

We first summarize where language-model features produced positive neural prediction (Figure~\ref{fig:positive_information}). We then evaluate how this signal changes under matched controls, participant-level summaries, response-profile tests, feature-ablation diagnostics, and calibration checks. This ordering keeps the positive predictive result as the empirical anchor and separates it from stronger evidence for model-specific predictive advantage, response-profile organization, computation-specific sensitivity, and reliability-bounded response-profile magnitude. The source tables preserve experimental coverage across models, layers, datasets, controls, response-profile metrics, feature-ablation diagnostics, reliability summaries, and positive controls; participant-level inference is reported for contrasts that retained participant or participant-run identifiers.

\FloatBarrier

\begin{figure}[!htbp]
\begin{center}
\includegraphics[width=0.96\textwidth]{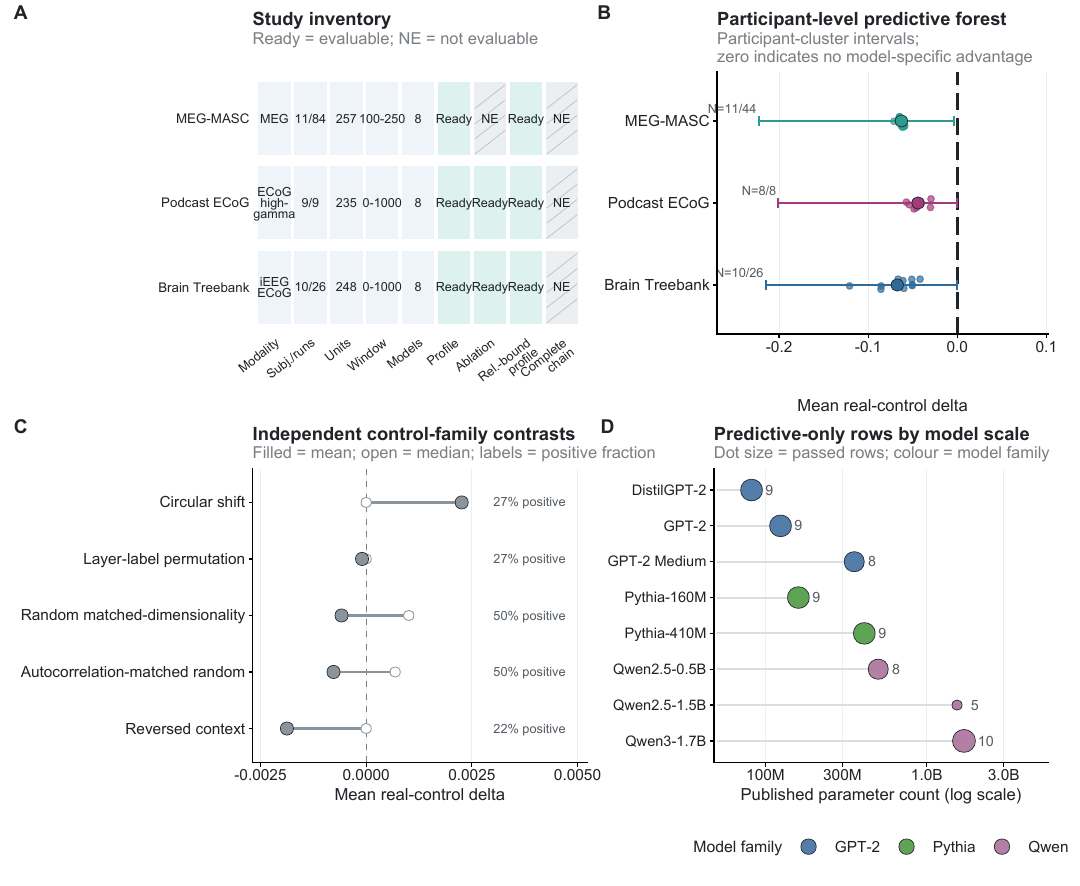}
\caption{Predictive coverage and matched controls. (A) Dataset inventory and branch availability in the matched derived data; Ready denotes branch coverage, including reliability-bounded profile rows where available, and NE denotes unavailable complete-chain coverage. (B) Participant-level model-control means summarize dataset-level scope; labels give participants/subject-run units retained after complete matching. (C) Independent control-family contrasts show mean Pearson-\(r\) real-minus-control deltas before the most-competitive-control reduction; point size gives the positive-row fraction, and the circular-shift contrast is slightly positive. (D) Model parameter scale and predictive-only row counts show passed rows across GPT-2, Pythia, and Qwen families, with no single family dominating.}
\label{fig:predictive_heterogeneity}
\end{center}
\end{figure}
\FloatBarrier

\subsection{Language-model features show heterogeneous neural predictivity}

Across the three primary datasets, language-model features produced widespread positive held-out prediction. In the Pearson predictive-control source table, 5541 of 11232 rows had a positive raw model score, and 5819 of 11232 rows improved over the nuisance baseline before matched-control comparison was considered. These source-level counts describe the prevalence of information-bearing predictivity and are separate from participant-level inferential units. Under the controlled predictive-only criterion, which additionally required the configured matched-control contrast at the summary-row level, 67 of 432 evaluable predictive rows were retained: 38 of 144 in Brain Treebank and 29 of 144 in Podcast ECoG. MEG-MASC contributed no predictive-only summary row. Participant-level mean raw Pearson-\(r\) summaries were small after averaging runs within participant: Brain Treebank had a mean of \(0.0058\) with a bootstrap interval of \([-0.0026, 0.0141]\), MEG-MASC had a mean of \(-0.0001\) with interval \([-0.0025, 0.0026]\), and Podcast ECoG had a mean of \(0.0033\) with interval \([-0.0048, 0.0123]\). Participant-level nuisance-baseline gains were similarly bounded: \(-0.0270\) for Brain Treebank, \(0.0077\) for MEG-MASC, and \(-0.0144\) for Podcast ECoG, with all corresponding bootstrap intervals crossing zero. Information-bearing predictivity was widespread in source-level summaries and remained detectable in a subset of controlled configurations, with participant-level averages defining the population-level scope.

Other components showed the same lower-level pattern. In the all-metric response-profile table, 313411 of 539136 metric cells had positive raw model-to-brain profile similarity. In the model-side feature-ablation table, 2704 of 3348 source rows had positive raw ablation deltas. Finally, 195632 of 539136 ceiling-normalized response-profile metric cells reached a raw fraction of ceiling of at least 0.25 before the model--control delta was considered. Together, these counts show that language-model-derived quantities can annotate neural responses and produce positive local summaries. The subsequent contrasts specify how far those summaries travel toward representational and computational interpretation.

\subsection{Predictive information is heterogeneous across datasets and model configurations}

The 67 predictive-only rows were distributed across the analyzed model set. Qwen3-1.7B contributed 10 rows; DistilGPT-2, Pythia-160M, Pythia-410M, and GPT-2 each contributed 9; Qwen2.5-0.5B-Instruct and GPT-2 Medium each contributed 8; and Qwen2.5-1.5B-Instruct contributed 5. The layer distribution was similarly broad, with 26 rows in final-layer features, 23 in middle-layer features, and 18 in embedding-layer features. The retained rows were concentrated in two candidate quantities in the matched data, with 35 semantic transition rows and 32 context update rows. Figure~\ref{fig:positive_information}D--E shows the model, layer, candidate-quantity, and participant-consistency pattern; Figure~\ref{fig:predictive_heterogeneity}D summarizes row counts by model parameter scale and family; and Supplementary Table 32 provides the compact table. The controlled positives support localized configuration-level predictive information with limited participant consistency across datasets.

The contrast-status summaries in Figure~\ref{fig:positive_information}D use status labels only for evaluable rows. A passed cell means that at least one summary row met the specified contrast rule. Dataset-level population support is evaluated separately through participant-aware summaries. Unavailable cells mark missing matched coverage.

\subsection{Local predictive information is heterogeneous at the participant level}

The most-competitive-control comparison paired each real model summary with nuisance features and the strongest matched control available for the same row (Figure~\ref{fig:predictive_heterogeneity}). The leading control family varied across rows: circular-shift controls led in 1097 rows, autocorrelation-matched random controls in 1108 rows, random matched-dimensionality controls in 633 rows, circular-shifted language-model controls in 669 rows, and layer-label permutation controls in 566 rows, with remaining rows assigned to token-order, within-story-block, sentence-reset, or reversed-context controls. Family-specific contrasts and control-family-removal checks provide complementary views of the same matched-control question.

At the dataset level, the most-competitive matched-control predictive comparison localized the controlled positives rather than showing a uniform participant-level advantage. Brain Treebank had a mean participant-cluster predictive delta of \(-0.0673\) across 26 retained subject-run units from 10 participants, with a participant-cluster interval of \([-0.2147, -0.0008]\). MEG-MASC had a mean participant-cluster predictive delta of \(-0.0627\) across 44 subject-run units from 11 participants, with an interval of \([-0.2226, -0.0038]\). Podcast ECoG had a mean participant-level predictive delta of \(-0.0440\) across 8 subject-run units from 8 participants, with an interval of \([-0.2012, 0.0000]\). The distinction between local summary positives and dataset-level means is the main heterogeneity result. Because no smallest effect size of interest was prespecified, these results provide interval bounds; formal equivalence tests would require a prespecified effect threshold.

The independent family-specific contrasts in Figure~\ref{fig:predictive_heterogeneity}C were at or below zero for four of the five displayed control families: random matched-dimensionality, autocorrelation-matched random, layer-label permutation, and reversed-context controls. The circular-shift contrast was slightly positive (\(0.0023\)), with 8973 of 32832 rows positive; the participant-level most-competitive-control summaries stayed below zero. The modality and time-window summary yielded the same coverage-bounded pattern. For Pearson-\(r\) predictive rows, Brain Treebank used the broad 0--1000 ms ECoG summary and had a mean most-competitive-control delta of \(-0.0780\). Podcast ECoG used the broad 0--1000 ms ECoG summary and had a mean most-competitive-control delta of \(-0.0440\). MEG-MASC used the broad 100--250 ms MEG summary and had a mean Pearson delta of \(-0.0627\). These summaries document the available modality and time-window coverage; anatomical or latency-specific hypotheses require finer retained target strata.

\begin{figure}[!htbp]
\begin{center}
\includegraphics[width=0.84\textwidth]{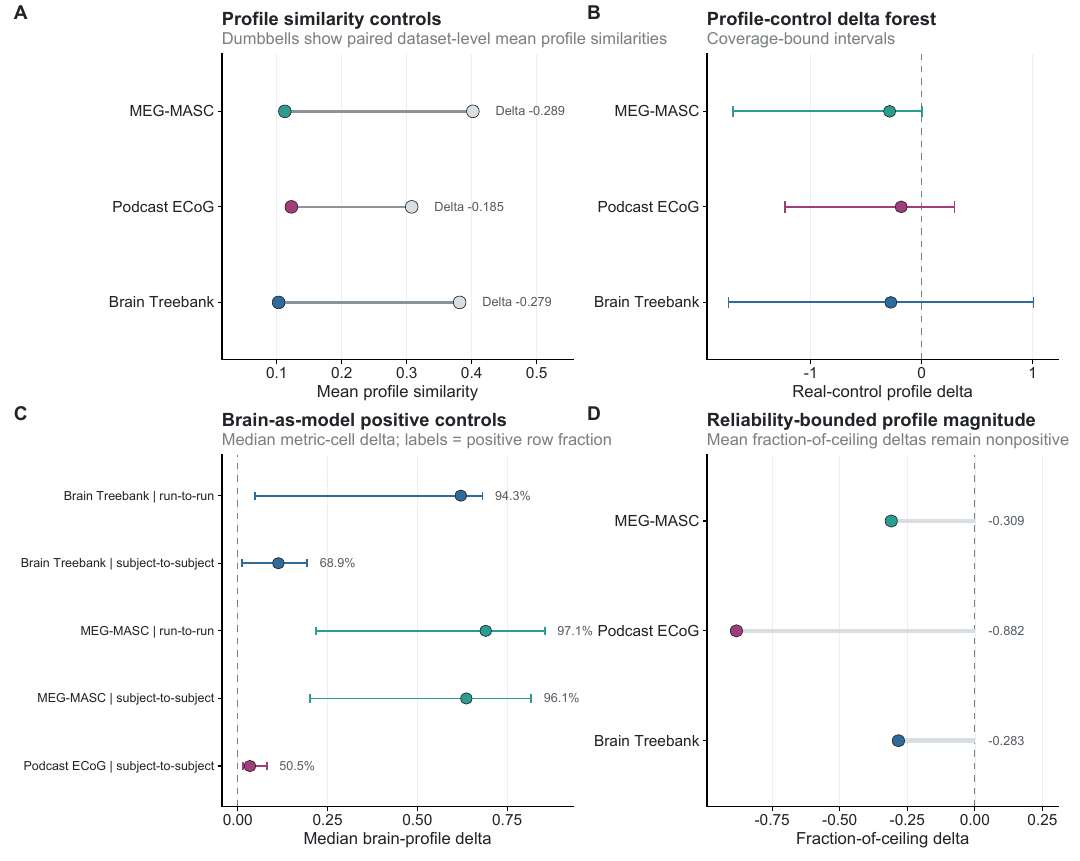}
\caption{Response-profile and reliability-bounded profile evidence. (A) Raw model-to-brain profile similarity is positive; best matched controls are stronger on average. (B) Dataset-level matched profile-control deltas are below the most competitive matched controls across all three primary datasets; intervals denote target-profile coverage summaries. (C) Brain-as-model positive controls show median metric-cell deltas versus shuffled unit order; bars show central metric-cell intervals and labels give positive row fractions. (D) Reliability-bounded fraction-of-ceiling deltas after matched profile controls.}
\label{fig:profile_reliability}
\end{center}
\end{figure}
\FloatBarrier

\subsection{Positive profile similarity localizes the evidence level}

The response-profile analysis asked whether model-to-brain profiles reproduce organization across sampled neural units beyond individual-target prediction. Raw model-to-brain profile similarities were positive in many metric cells, as noted above. Matched profile deltas were available for all three primary datasets in the matched derived data, and the dataset-level mean real-minus-best-control profile deltas were below the most competitive matched controls. Brain Treebank had a mean real model-to-brain profile similarity of \(0.1028\), a mean best-control similarity of \(0.3815\), and a mean delta of \(-0.2787\). MEG-MASC had corresponding values of \(0.1124\), \(0.4019\), and \(-0.2895\). Podcast ECoG had corresponding values of \(0.1225\), \(0.3077\), and \(-0.1852\).

This pattern localizes the evidence level. The tested representations produced positive raw profile similarities, whereas the matched profile-control contrast placed shared cross-neural-unit organization beyond the current dataset-mean evidence. Other relationships between language models and neural responses remain outside what this contrast tests.

\subsection{Model-side ablations reveal widespread sensitivity}

Model-side feature-ablation summaries were available for Brain Treebank and Podcast ECoG in the matched participant-run data. Brain Treebank had a mean raw feature-ablation delta of \(0.0948\), with 2133 of 2700 table-row deltas positive. Podcast ECoG had a mean raw feature-ablation delta of \(0.0967\), with 571 of 648 table-row deltas positive. These positive deltas show that candidate-quantity ablations can change held-out neural prediction. In the present evidence framework, selective computational correspondence is a higher evidence level that additionally requires predictive matched-control support, feature-specificity support, and aligned response-profile evidence.

The interpretation differs by candidate quantity. Surprisal remains useful as a predictive annotation of contextual predictability. Semantic transition deltas are diagnostic for semantic-state coding hypotheses. Dependency integration deltas are diagnostic for syntactic or combinatorial hypotheses. Boundary-related deltas are diagnostic for localized boundary-detector interpretations. Context update deltas are diagnostic for shared model--brain context-accumulation interpretations. These outcomes support diagnostic sensitivity, with strong computational correspondence requiring evidence beyond the current feature-ablation summaries.

\subsection{Reliability ceilings bound response-profile magnitude}

Reliability normalization calibrated the profile signal against reliable brain-derived profile structure. After aggregation to the ceiling table, the brain--brain reliability summaries contained 36 dataset-by-method rows, of which 18 met the configured 0.10 reliability minimum. Ceiling-normalized response-profile rows were available for all three primary datasets. The mean fraction-of-ceiling deltas were \(-0.2829\) for Brain Treebank, \(-0.3093\) for MEG-MASC, and \(-0.8823\) for Podcast ECoG after matched profile controls. These results locate the positive raw profile similarities below the reliability-bounded evidence level and keep them separate from predictive and feature-ablation contrasts.

\subsection{Brain-derived and implanted signals validate component-level sensitivity}

The control analyses tested whether the analysis could recover signal in settings with expected structure. The most complete integrated check was the implanted-signal simulation. Its single strong-signal row produced a predictive score of \(0.9867\), a predictive delta of \(0.8945\), a response-profile delta of \(1.6154\), a model-side feature-ablation delta of \(0.2659\), and a fraction of ceiling of \(0.9911\). A stochastic graded version of this synthetic check is shown in Figure~\ref{fig:ablation_calibration}C. Repeating each candidate-signal strength across 100 seeded random draws yielded an interpolated 80\% integrated-detection threshold of 1.49. The strength axis is the synthetic coefficient \(\beta\) in the engineered latent model; 1.49 is therefore an implementation-scale detection threshold outside empirical neural effect-size units. This engineered control shows recovery of a known signal once it is strong enough in the synthetic setting, with real-data sensitivity bounded by the available participant-run and derived target units.

The brain-derived controls support component-level sensitivity (Figure~\ref{fig:profile_reliability}C). Across primary and secondary reliability checks, all 111 valid brain--brain reliability rows exceeded the configured minimum reliability of 0.10. After aggregation by dataset and method, 18 of 36 reliability summaries met the same threshold. The brain-as-model panel reports median deltas against shuffled brain-unit order. Brain Treebank showed positive run-to-run and subject-to-subject median deltas of \(0.621\) and \(0.113\), with positive-row fractions of 94.3\% and 68.9\%. MEG-MASC showed positive run-to-run and subject-to-subject median deltas of \(0.690\) and \(0.636\), with positive-row fractions of 97.1\% and 96.1\%. Podcast ECoG subject-to-subject brain-as-model profiles were weaker but positive and are interpreted as auxiliary: the median delta was \(0.034\), with 50.5\% positive rows.

Low-level and acoustic checks showed source-limited sensitivity to timing-linked signals. Brain Treebank word-onset checks exceeded the configured criterion in 5 of 30 table rows in the 0--100 ms window and 4 of 30 table rows in the 100--300 ms window. MEG-MASC word-onset and word-rate checks also produced partial evidence. WAV-derived acoustic-envelope readouts were available for MEG-MASC local audio and Podcast ECoG standalone and full-length checks. The full-length Podcast acoustic check exceeded the configured criterion in 6 of 8 readable subject-run rows, with median delta over the best control of \(0.005433\). Brain Treebank is retained as a source-limited case for acoustic-envelope analysis because rights-cleared waveform or movie source files were unavailable.

\subsection{Robustness analyses preserve the same pattern}

\begin{figure}[!htbp]
\begin{center}
\includegraphics[width=0.92\textwidth]{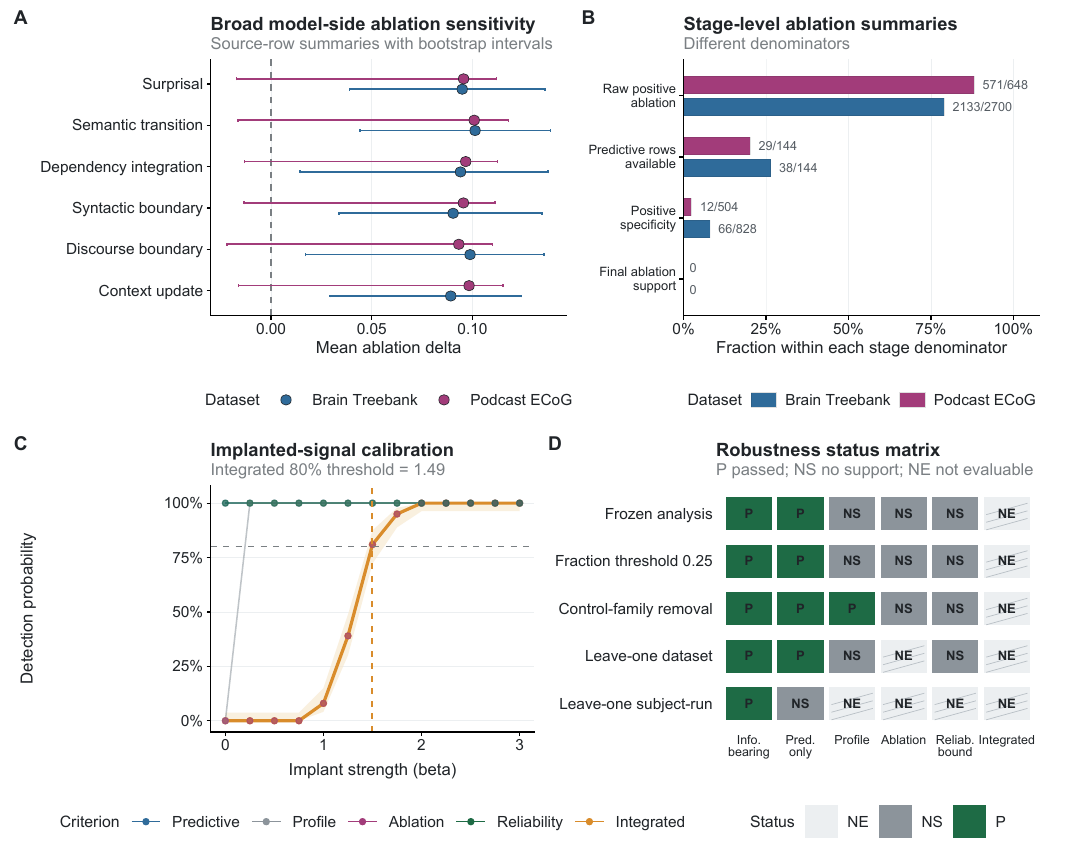}
\caption{Feature-ablation, specificity, and calibration. (A) Model-side ablation deltas for Brain Treebank and Podcast ECoG candidate quantities. (B) Stage-level summaries with distinct denominators. (C) Stochastic implanted-signal calibration with 100 seeded implants per strength; the band gives Wilson 95\% intervals, and dashed lines mark 80\% detection and the threshold of 1.49. (D) Robustness status matrix; P denotes at least one passed row, NS evaluated without support, and NE unavailable matched coverage.}
\label{fig:ablation_calibration}
\end{center}
\end{figure}

Robustness analyses summarize how the results change across evidence levels (Figure~\ref{fig:ablation_calibration}D). The evaluable predictive summary table contains 432 rows: 144 rows in each primary dataset. These rows are coverage summaries over tested model--dataset--layer combinations, separate from participant counts and complete-chain tests. In the matched derived data, response-profile rows were matched for all three datasets, participant-run feature-ablation rows were matched for Brain Treebank and Podcast ECoG, and remaining unavailable contrasts lacked matched coverage. Thirty-eight Brain Treebank rows and 29 Podcast ECoG rows were labeled predictive-only by the configured decision rule. Complete co-indexing across all required contrasts remains a future coverage requirement, so Figure~\ref{fig:ablation_calibration}D marks those chains as NE.

Leave-one-dataset-out checks, reliability-threshold sweeps, fraction-of-ceiling sweeps, positive-direction sweeps, most-competitive-control aggregation sweeps, and single-control-family removals preserved the same qualitative pattern for real language-model rows. Some relaxed rules created stage-level positives by removing or weakening a relevant contrast. Jointly indexed predictive, response-profile, feature-ablation, reliability-bounded, matched-control, and replication evidence for the same row remains a future coverage requirement. Information-bearing predictivity and diagnostic ablation sensitivity were reproducible across these checks, while response-profile organization and candidate-computation correspondence occupied higher evidence levels under the specified contrasts.

\FloatBarrier

\section{Discussion}

\subsection{Heterogeneous predictivity during naturalistic comprehension}

Language-model features provided heterogeneous, information-bearing neural predictivity during naturalistic comprehension. Positive held-out prediction and nuisance-baseline gains were widespread, a subset of controlled configurations met the predictive-only criterion, and model-side ablations frequently altered neural prediction. These results show where language-model features act as useful neural annotations and what additional evidence is needed for model-specific, representational, or computation-specific interpretations.

These results are tied to the analyzed datasets and fixed representation files. The analysis therefore treats predictive usefulness, response-profile correspondence, candidate-computation specificity, and reliability-bounded response-profile magnitude as different claims. A positive local encoding score supports predictive usefulness; stronger statements require additional matched-control, profile, ablation-specific, and reliability-bounded evidence.

\subsection{Local information and shared organization are empirically dissociable}

Naturalistic language contains structure at many time scales. Word onsets, word rate, acoustic envelope, sentence position, discourse progression, lexical frequency, token predictability, and temporal autocorrelation can all help a flexible readout predict neural measurements. A high-dimensional language-model representation may capture some of this structure while leaving cross-unit organization unresolved. Matched controls and response-profile contrasts clarify what kind of theoretical conclusion the data can support.

The response-profile results were especially informative. The tested models had positive raw profile similarities in many rows, and the matched profile deltas were lower than the most competitive matched control in all three primary datasets. This pattern is consistent with predictive or annotational usefulness and locates shared cross-neural-unit response-profile organization at a higher evidence level than the current contrasts support.

\subsection{Ablation sensitivity provides diagnostic evidence about candidate quantities}

The feature-ablation results sharpen model-based accounts of language comprehension. Surprisal remains a useful candidate annotation of contextual predictability, and prior work links surprisal to neural responses during natural speech and reading \citep{weissbart2020surprisal,michaelov2024surprisal}. In the present analysis, surprisal-related ablation effects indicated sensitivity of fitted neural readouts to model-side information removal. Selective dependence on the tested language-model surprisal computation would require additional support from the predictive matched-control contrast and feature-specificity diagnostics.

The same logic applies to the other tested quantities. Semantic transition effects remain diagnostic for semantic-state coding hypotheses. Dependency integration effects remain diagnostic for syntactic or combinatorial hypotheses. Syntactic and discourse boundary effects remain diagnostic for localized boundary-detector interpretations. Context update effects remain diagnostic for shared model--brain context-accumulation interpretations. These quantities remain theoretically meaningful, and strong computational correspondence requires evidence beyond raw or diagnostic ablation sensitivity.

\subsection{Implications for brain--AI model evaluation}

The findings are compatible with prior reports that language-model features predict neural responses during comprehension \citep{schrimpf2021,goldstein2022,caucheteux2022,hosseini2024nol}. Those studies often ask whether model-derived quantities contain information useful for explaining neural or behavioral measurements. The present analyses add a complementary inferential question: whether the same comparisons support claims about response-profile organization or candidate language computations after matched controls, source-coverage checks, ablation diagnostics, and reliability-bounded response-profile interpretation.

Future brain--AI model evaluation should separately report positive held-out prediction, nuisance-baseline gain, matched-control advantage, response-profile organization, computation-specificity diagnostics, and reliability-bounded response-profile magnitude. This separation preserves the scientific value of language-model features as neural annotations and keeps model-specific representational claims tied to the evidence required to support them.

\subsection{Relation to prior control analyses}

Hadidi and colleagues establish that methodological choices and stimulus-related variables can inflate neural predictivity, and that positional signals and word rate can perform competitively with trained language models in widely used datasets \citep{hadidi2026spurious}. The present findings build on this concern while retaining the observed information-bearing predictivity of model features. Substantial information-bearing predictivity and localized predictive-only positives appeared alongside separate tests of response-profile organization and computation-specific correspondence.

The two studies therefore address different inferential transitions: robustness of prediction and interpretation of surviving prediction. The analyses here separate predictive information, cross-neural-unit response-profile correspondence, candidate-computation ablation sensitivity, and reliability-bounded response-profile magnitude.

\subsection{Limitations and future tests}

Several scope limits remain. The analysis relies on a fixed derived-data set and contains no new prospective data collection. Dataset coverage is uneven across evidence levels. Participant-run predictive summaries cover 10 Brain Treebank participants and 26 subject-run units, 11 MEG-MASC participants and 44 subject-run units, and 8 Podcast ECoG participants and 8 subject-run units, which are still modest samples for participant-cluster inference. Response-profile deltas and ceiling-normalized profile summaries use target-profile grids, while feature-ablation closure is currently available for Brain Treebank and Podcast ECoG through matchable derived grids. These contrast-specific scopes define coverage limits without population-equivalence estimates.

The positive controls also have contrast-specific scope. The implanted-signal calibration in Figure~\ref{fig:ablation_calibration}C repeats stochastic synthetic implants across candidate-signal strengths and estimates the strength needed for high-probability detection. This engineered simulation calibrates the implementation-level detection path; participant-level detection power over real participants, regions, or stimulus samples remains unresolved.

The model inventory is bounded to the fixed representation files. Larger open models, paired base and instruction-tuned models, and prospective holdout applications could change the evidence landscape. Future work should also broaden MEG-MASC contrast-level participant-run coverage and response-profile coverage, prespecify a smallest effect size of interest before formal equivalence testing, add finer anatomical and latency-specific analyses, and estimate participant-level detection sensitivity where source data permit those analyses.

\subsection{Conclusion}

This study identifies heterogeneous neural predictivity from language models during naturalistic comprehension. Across three datasets, language-model features produced widespread positive held-out prediction, nuisance-baseline gains, localized predictive-only effects, and sensitivity to model-side ablation. These results support language-model features as informative neural annotations in the analyzed derived data. The same analyses define the additional conditions needed for claims about model-specific advantage, shared cross-neural-unit response organization, and shared language-processing computations.

\section*{Data and Code Availability Statements}
Derived result tables, analysis code, manuscript tables, and reproducibility scripts are available through an OSF view-only repository: \url{https://osf.io/7s84h/overview?view_only=e4a83f8cece44c90af40417820c8acd8}. Raw neural datasets and stimulus media are not redistributed and remain governed by their original data providers. Upon acceptance, the repository record will be made public and assigned or updated with a persistent accession or DOI.

\section*{AI-Assisted Tools Disclosure}

AI-assisted tools were used during manuscript preparation only for manuscript grammar and clarity checks and for code-quality review. The author reviewed and verified all final text, analyses, interpretations, citations, code, and submission materials, and accepts full responsibility for the content.

\section*{Supplementary Material}

Supplementary Information follows the references in this arXiv source package.

\printbibliography[heading=bibintoc]

\clearpage
\appendix
\section*{Supplementary Information}
\addcontentsline{toc}{section}{Supplementary Information}
\setcounter{figure}{0}
\setcounter{table}{0}
\renewcommand{\thefigure}{S\arabic{figure}}
\renewcommand{\thetable}{S\arabic{table}}

\section*{Supplementary Overview}

This Supplementary Information documents the reproducibility and coverage materials supporting the manuscript. The main manuscript evaluates what cognitive-neuroscientific inferences language-model neural predictivity supports during naturalistic language comprehension. This supplement provides the table index, source-coverage notes, criterion implementation, and supplementary figures needed to interpret the manuscript.

The supplement should be read together with the derived CSV tables available through the OSF data-and-code repository. Raw neural data and stimulus media are not redistributed. Terminology matches the main manuscript: predictive, response-profile, feature-ablation, reliability-bounded profile, and integrated evidence levels.

\section*{Supplementary Figures}

\begin{figure}[p]
\centering
\includegraphics[width=\textwidth]{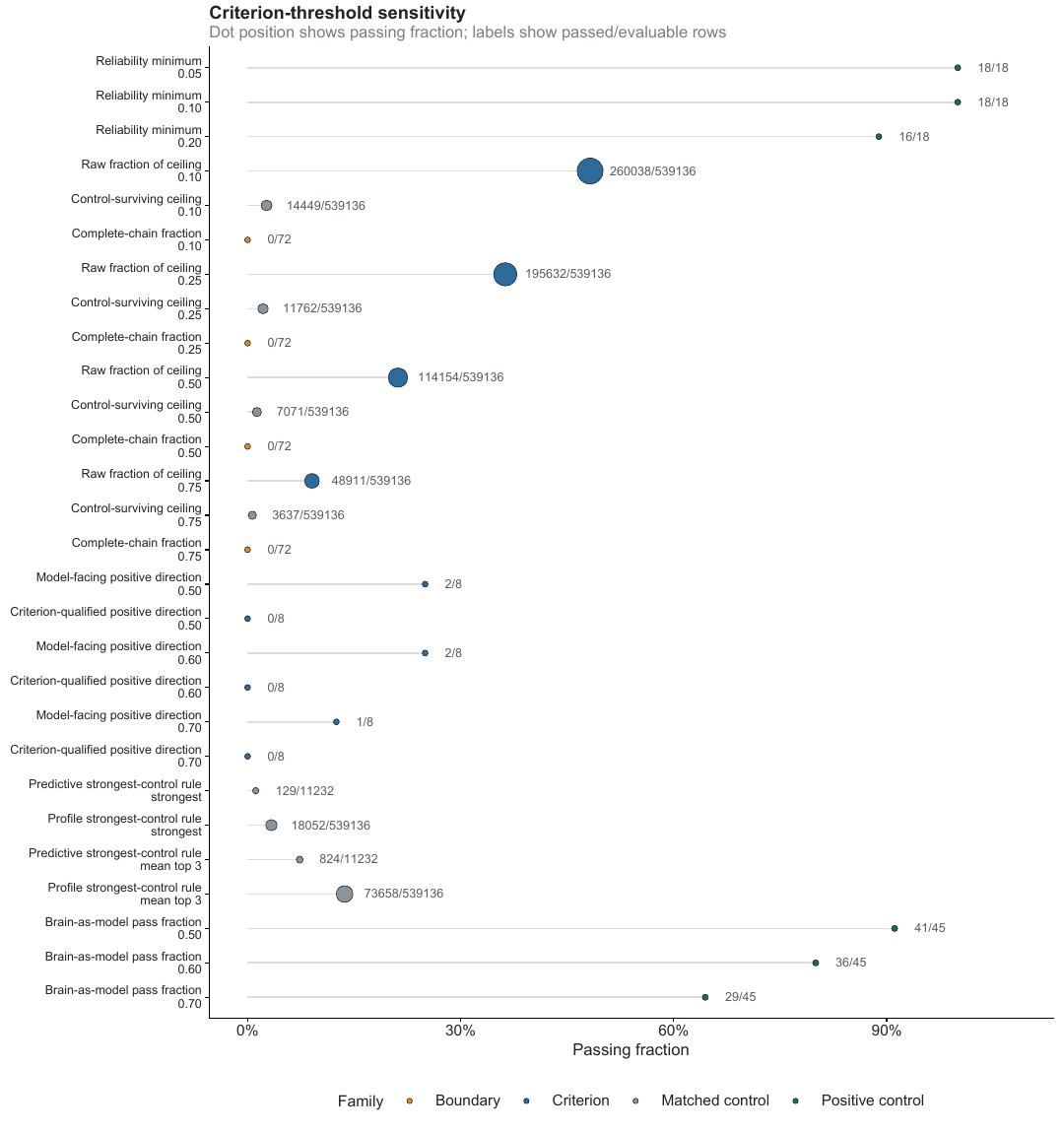}
\caption{Criterion-threshold sensitivity matrix. Threshold and rule sweeps summarize criterion status under relaxed rules; component-level positives are reported separately from jointly indexed integrated coverage.}
\end{figure}
\clearpage

\begin{figure}[p]
\centering
\includegraphics[width=\textwidth]{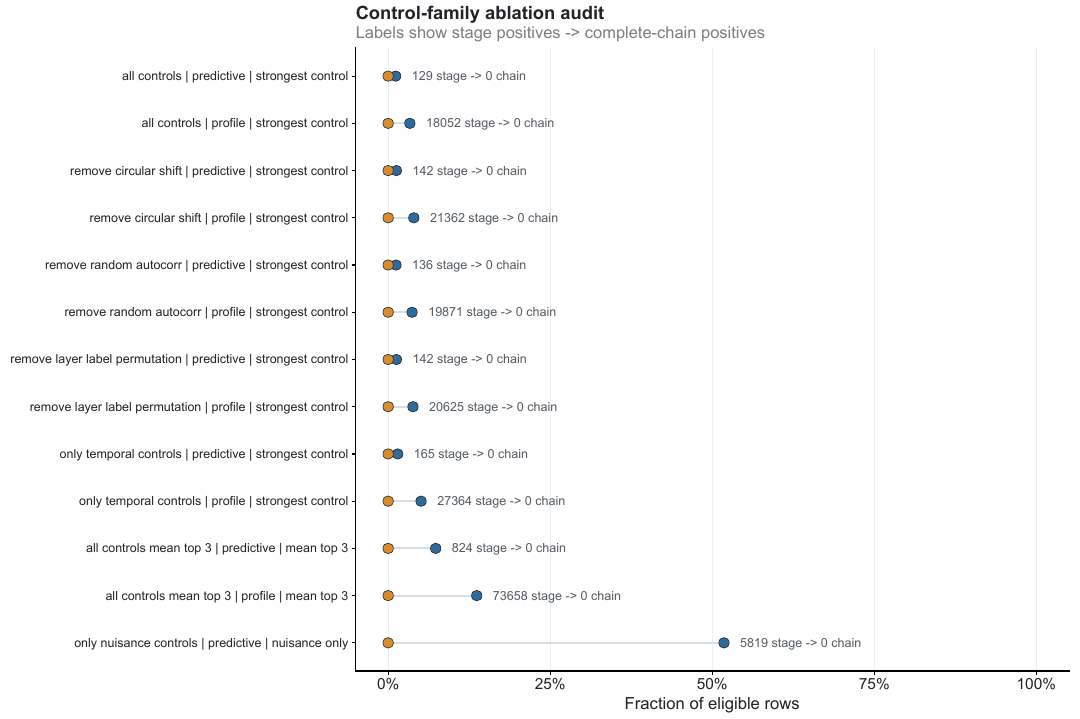}
\caption{Control-family ablation. Labels report stage positives to complete-chain positives. Single-family removal leaves the integrated interpretation outside jointly indexed complete-chain coverage.}
\end{figure}
\clearpage

\begin{figure}[p]
\centering
\includegraphics[width=\textwidth]{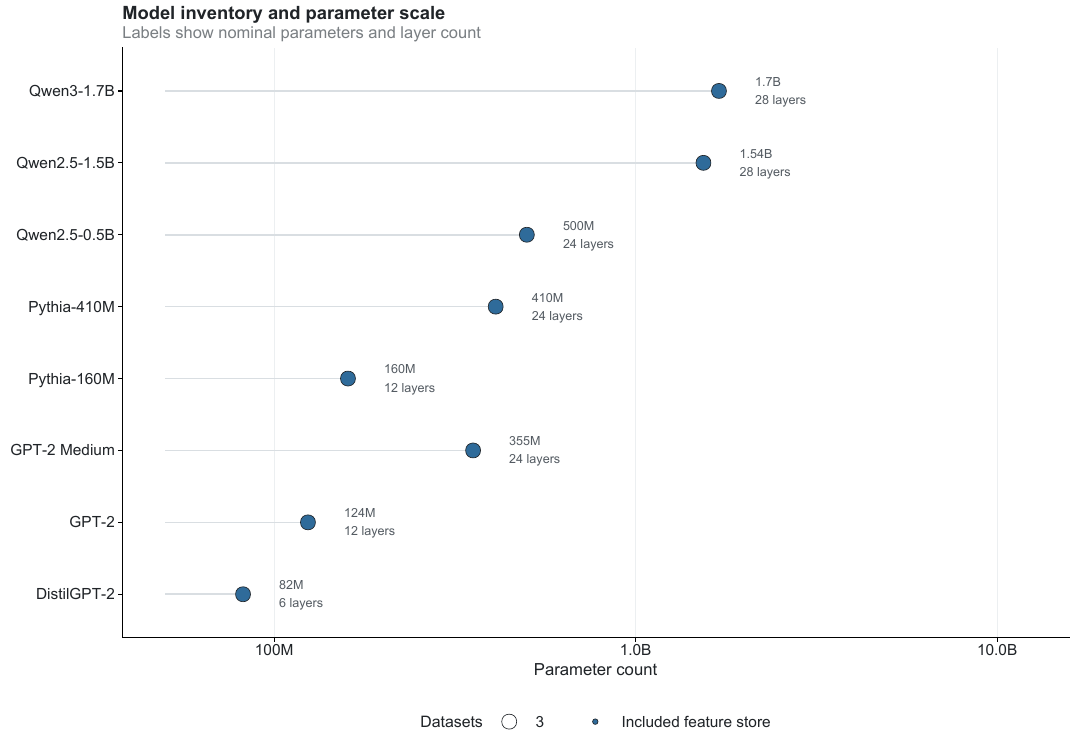}
\caption{Model inventory and coverage limits. The included-feature-store label identifies models retained in the fixed representation-file set. The final summary table is bounded to fixed representation files and matched analysis rows.}
\end{figure}
\clearpage

\begin{figure}[p]
\centering
\includegraphics[width=\textwidth]{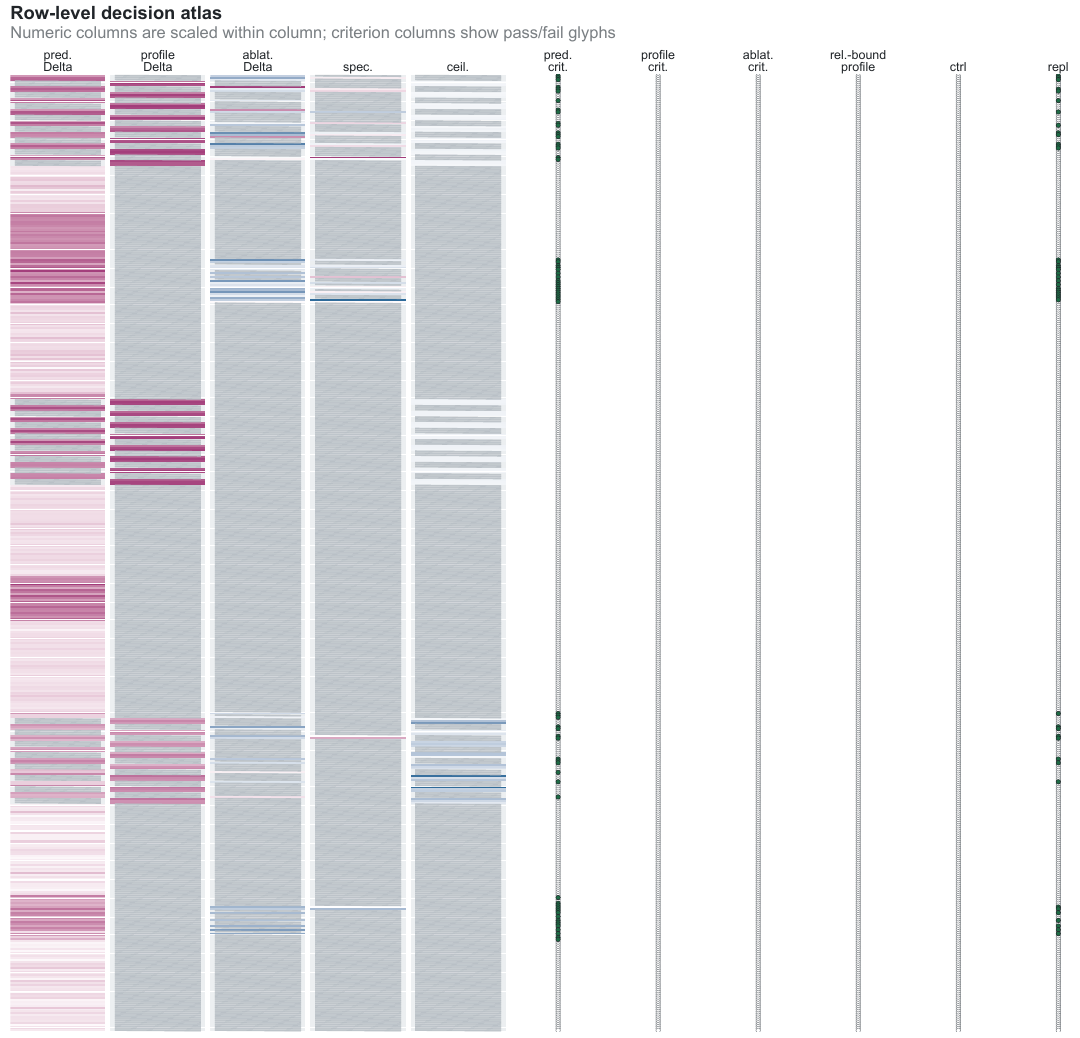}
\caption{Integrated-coverage heatmap. The integrated table retains 504 real-model coverage rows, of which 432 are evaluable predictive summary rows in the matched derived data.}
\end{figure}
\clearpage

\begin{figure}[p]
\centering
\includegraphics[width=\textwidth]{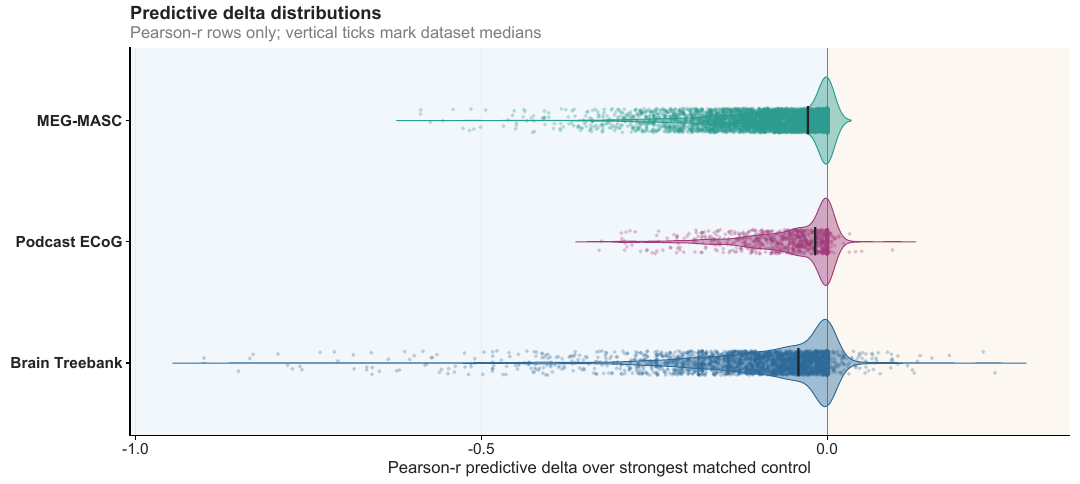}
\caption{Predictive delta distributions by dataset and control family. Dataset-level mean predictive contrasts are nonpositive, although a subset of evaluable summary rows meets the predictive-only criterion.}
\end{figure}
\clearpage

\begin{figure}[p]
\centering
\includegraphics[width=\textwidth]{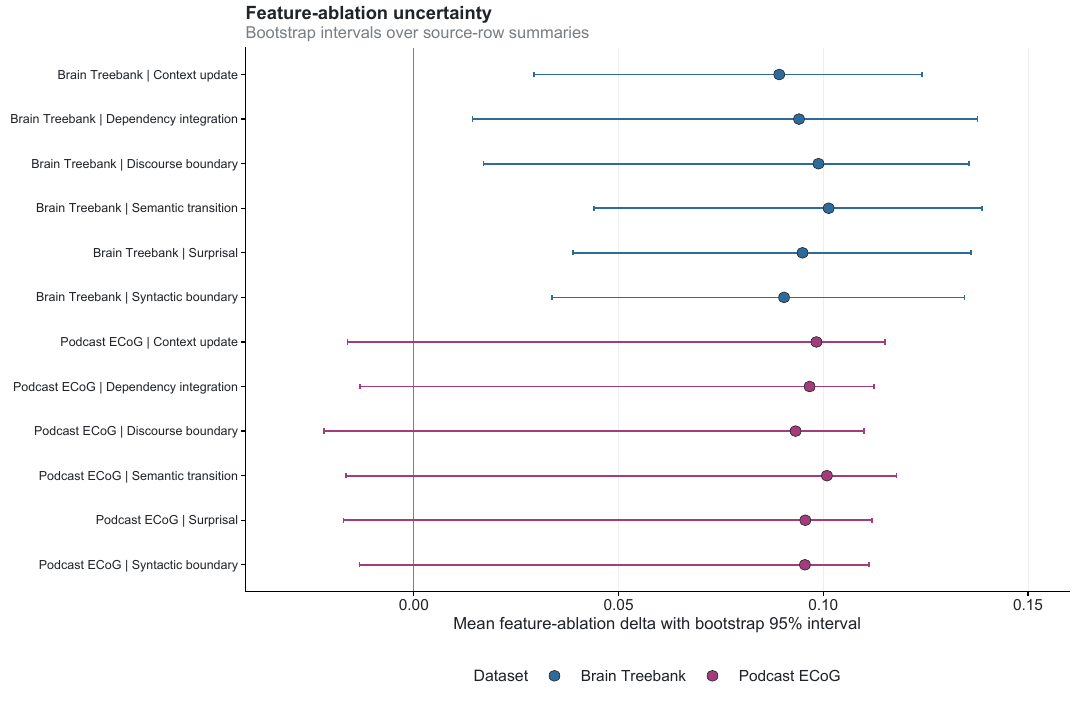}
\caption{Model-side feature-ablation uncertainty by candidate quantity. Intervals are bootstrap intervals over source-row summaries, not participant-level population intervals. Positive ablation deltas are diagnostic-level evidence; ablation-supported computational correspondence requires predictive matched-control evidence, feature-specificity diagnostics, and response-profile evidence for the integrated claim.}
\end{figure}
\clearpage

\begin{figure}[p]
\centering
\includegraphics[width=0.86\textwidth,height=0.64\textheight,keepaspectratio]{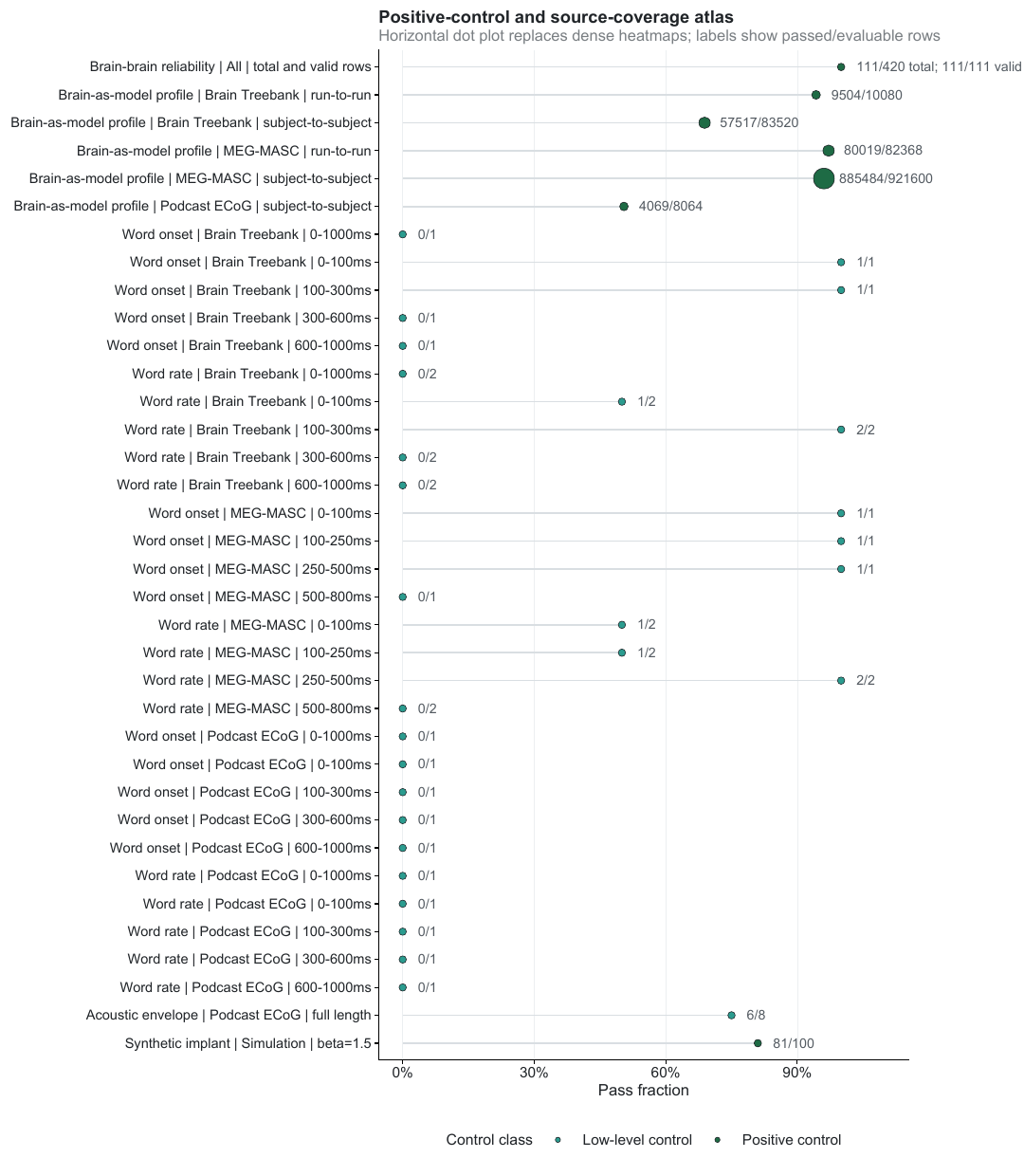}
\caption{Positive-control and source-coverage atlas. The horizontal dot plot collects brain--brain reliability, brain-as-model profile controls, word-onset and word-rate checks, full-length Podcast acoustic-envelope checks, and the stochastic implanted-signal calibration. The brain--brain reliability row reports both total table rows and valid-row coverage; other labels use the denominators of the corresponding source tables. The atlas makes auxiliary checks and source-coverage boundaries visible; component-level positives are interpreted at their corresponding evidence level.}
\end{figure}

\clearpage
\section*{Supplementary Table S0: Figure Reproducibility Index}

Supplementary Table S0 is provided as a machine-readable CSV at:

\begin{verbatim}
anc/figure_source_data/
Supplementary_Table_S0_figure_reproducibility_index.csv
\end{verbatim}

The table maps each main-figure and atlas panel to its manuscript-facing source-data CSV, source tables, generation script, seed policy, input directory, output files, and MD5 checksums for the source-data and output files. This panel-level index connects the visual claims in Figures 1--4 and Supplementary Figure S7 to the exact derived tables and R outputs.

\begingroup
\footnotesize
\setlength{\tabcolsep}{3pt}
\renewcommand{\arraystretch}{1.12}
\noindent\begin{tabularx}{0.90\linewidth}{p{0.08\linewidth}p{0.06\linewidth}Yp{0.29\linewidth}}
\toprule
Figure & Panel & Panel role & Source-data export \\
\midrule
Fig. 1 & A & Analysis schematic & \texttt{Fig1\_source\_data.csv} \\
Fig. 1 & B & Participant-level raw held-out prediction & \texttt{Fig1\_source\_data.csv} \\
Fig. 1 & C & Participant-level nuisance-baseline gain & \texttt{Fig1\_source\_data.csv} \\
Fig. 1 & D & Predictive-only model-by-dataset-by-layer heatmap & \texttt{Fig1\_source\_data.csv} \\
Fig. 1 & E & Participant consistency for predictive-only configurations & \texttt{Fig1\_source\_data.csv} \\
Fig. 2 & A & Study inventory and branch availability & \texttt{Fig2\_source\_data.csv} \\
Fig. 2 & B & Participant-level model-control means & \texttt{Fig2\_source\_data.csv} \\
Fig. 2 & C & Independent control-family contrasts & \texttt{Fig2\_source\_data.csv} \\
Fig. 2 & D & Predictive-only positives by model parameter scale and family & \texttt{Fig2\_source\_data.csv} \\
Fig. 3 & A & Raw profile similarity versus matched controls & \texttt{Fig3\_source\_data.csv} \\
Fig. 3 & B & Profile-control delta & \texttt{Fig3\_source\_data.csv} \\
Fig. 3 & C & Brain-as-model positive control & \texttt{Fig3\_source\_data.csv} \\
Fig. 3 & D & Reliability-bounded profile magnitude & \texttt{Fig3\_source\_data.csv} \\
Fig. 4 & A & Raw feature-ablation delta by candidate quantity & \texttt{Fig4\_source\_data.csv} \\
Fig. 4 & B & Stage-level feature-ablation summaries with distinct denominators & \texttt{Fig4\_source\_data.csv} \\
Fig. 4 & C & Stochastic implanted-signal detection curve & \texttt{Fig4\_source\_data.csv} \\
Fig. 4 & D & Threshold/control/dataset robustness perturbations across evidence levels, including reliability-bound profile magnitude & \texttt{Fig4\_source\_data.csv} \\
Fig. S7 & all & Positive-control and source-coverage atlas & \texttt{FigS7\_source\_data.csv} \\
\bottomrule
\end{tabularx}
\endgroup

\newpage
\section*{Key Supplementary Tables Displayed in This PDF}

The detailed machine-readable CSV files remain the authoritative table exports. The compact tables below reproduce the submission-facing content needed to read the Methods and Results alongside the source files.

\subsection*{Supplementary Table 3. Binary criterion definitions, alternatives addressed, and interpretation}

\begingroup
\scriptsize
\setlength{\tabcolsep}{2pt}
\renewcommand{\arraystretch}{0.95}
\noindent\begin{tabularx}{0.98\linewidth}{p{0.17\linewidth}p{0.30\linewidth}p{0.21\linewidth}Y}
\toprule
Criterion & Required evidence & Main alternative addressed & Interpretation in this manuscript \\
\midrule
Predictive & Real model exceeds nuisance and matched controls in held-out prediction. & Timing, lexical statistics, temporal autocorrelation, and capacity. & Sixty-seven of 432 evaluable predictive summary rows pass the predictive-only rule: 38/144 in Brain Treebank and 29/144 in Podcast ECoG. Participant-level mean contrasts are nonpositive in all three primary datasets. \\
Response profile & Model-to-brain profiles exceed shuffled or matched profile controls over sampled neural units after summary-row aggregation. & Marginal prediction or isolated positive target rows without cross-unit organization. & Dataset-level profile-control deltas are nonpositive, and aggregate response-profile summary rows do not meet the criterion. \\
Feature ablation & Full model exceeds model-side ablated representation with feature-specificity diagnostics, conditional on predictive evidence. & Generic feature loss or nonspecific dimensional change. & Positive deltas are diagnostic only. \\
Reliability-bounded response-profile magnitude & A surviving aggregate response-profile effect is interpreted relative to a valid brain--brain profile reliability estimate. & Low neural reliability or ceiling limitations. & Raw response-profile fractions can be positive; control-surviving aggregate rows do not meet the criterion. \\
Integrated & Predictive, response-profile, feature-ablation, reliability-bounded response-profile magnitude, matched-control, and replication criteria all hold. & Single-contrast or source-limited positives. & The available data lacked a jointly indexed complete chain across all required contrasts; unavailable contrasts mark missing matched coverage. \\
\bottomrule
\end{tabularx}
\endgroup

\subsection*{Supplementary Table 4. Matched-control families}

\begingroup
\scriptsize
\setlength{\tabcolsep}{2pt}
\renewcommand{\arraystretch}{0.95}
\noindent\begin{tabularx}{0.98\linewidth}{p{0.24\linewidth}p{0.26\linewidth}Y}
\toprule
Control family & Contrast role & Alternative explanation addressed \\
\midrule
Nuisance baseline & Low-level covariate comparator. & Word timing, lexical covariates, and non-model features. \\
Random matched-dimensionality & Strict matched-capacity control. & High-dimensional readout flexibility. \\
Autocorrelation-matched random & Strict temporal-structure control. & Temporal autocorrelation in neural and stimulus streams. \\
Circular shift & Strict temporal misalignment control. & Slow trends and nonlocal temporal structure. \\
Sentence reset & Context-history control. & Benefits from long context beyond local token content. \\
Reversed context & Context-direction control. & Contextual history not tied to natural forward order. \\
Layer-label permutation & Layer-specificity control. & Layer effects attributable to label or indexing artifacts. \\
Token-order shuffle & Word-order control. & Lexical content without natural order. \\
Within-story block shuffle & Local block-order control. & Story-local structure not tied to exact event order. \\
\bottomrule
\end{tabularx}
\endgroup

\newpage
\begin{landscape}
\subsection*{Supplementary Tables 17--18. Dataset-level uncertainty and interval-bound audit}

\begingroup
\fontsize{5.8}{6.3}\selectfont
\setlength{\tabcolsep}{1.8pt}
\renewcommand{\arraystretch}{0.90}
\noindent\begin{tabularx}{0.98\linewidth}{p{0.15\linewidth}p{0.22\linewidth}p{0.07\linewidth}p{0.12\linewidth}p{0.11\linewidth}p{0.14\linewidth}Y}
\toprule
Dataset & Contrast & Rows & Participants / units & Mean delta & Interval bound & Interpret. \\
\midrule
Brain Treebank & Predictive model-control delta & 144 & 10 / 26 & -0.0673 & [-0.2147, -0.0008] & Participant-cluster summary; no formal equivalence test without SESOI. \\
MEG-MASC & Predictive model-control delta & 144 & 11 / 44 & -0.0627 & [-0.2226, -0.0038] & Participant-cluster summary; no formal equivalence test without SESOI. \\
Podcast ECoG & Predictive model-control delta & 144 & 8 / 8 & -0.0440 & [-0.2012, 0.0000] & Participant-level predictive summary; no formal equivalence test without SESOI. \\
Brain Treebank & Response-profile model-control delta & 179712 & 10 / 26 & -0.2787 & [-1.7413, 1.0070] & Source-target/profile summary; outside population-equivalence testing. \\
MEG-MASC & Response-profile model-control delta & 304128 & 11 / 44 & -0.2895 & [-1.6988, 0.0000] & Source-target/profile summary; outside population-equivalence testing. \\
Podcast ECoG & Response-profile model-control delta & 55296 & 8 / 8 & -0.1852 & [-1.2348, 0.2945] & Source-target/profile summary; outside population-equivalence testing. \\
Brain Treebank & Feature-ablation delta & 2700 & 10 / 15 & 0.0948 & [-0.4109, 0.5321] & Diagnostic summary below integrated support. \\
Podcast ECoG & Feature-ablation delta & 648 & 5 / 5 & 0.0967 & [-0.1003, 0.2455] & Diagnostic summary below integrated support. \\
\bottomrule
\end{tabularx}
\endgroup
\end{landscape}

\newpage
\subsection*{Supplementary Table 26. Model inventory summary}

\begingroup
\fontsize{6.8}{7.6}\selectfont
\setlength{\tabcolsep}{1.8pt}
\renewcommand{\arraystretch}{0.94}
\noindent\begin{tabularx}{0.98\linewidth}{p{0.22\linewidth}p{0.08\linewidth}p{0.10\linewidth}p{0.06\linewidth}p{0.20\linewidth}p{0.08\linewidth}Y}
\toprule
Model & Family & Published parameters & Layers & Datasets used & Decision rows & Status \\
\midrule
Pythia-160M & Pythia & 160M & 12 & Brain Treebank; MEG-MASC; Podcast ECoG & 63 & Included \\
Pythia-410M & Pythia & 410M & 24 & Brain Treebank; MEG-MASC; Podcast ECoG & 63 & Included \\
Qwen2.5-0.5B-Instruct & Qwen & 0.5B & 24 & Brain Treebank; MEG-MASC; Podcast ECoG & 63 & Included \\
Qwen2.5-1.5B-Instruct & Qwen & 1.54B & 28 & Brain Treebank; MEG-MASC; Podcast ECoG & 63 & Included \\
Qwen3-1.7B & Qwen & 1.7B & 28 & Brain Treebank; MEG-MASC; Podcast ECoG & 63 & Included \\
DistilGPT-2 & GPT-2 & 82M & 6 & Brain Treebank; MEG-MASC; Podcast ECoG & 63 & Included \\
GPT-2 & GPT-2 & 124M & 12 & Brain Treebank; MEG-MASC; Podcast ECoG & 63 & Included \\
GPT-2 Medium & GPT-2 & 355M & 24 & Brain Treebank; MEG-MASC; Podcast ECoG & 63 & Included \\
Qwen2.5-7B-Instruct & Qwen & 7.6B & 28 & none & 0 & Exceeds cap \\
Qwen3-4B-Instruct-2507 & Qwen & 4.0B & 36 & none & 0 & Exceeds cap \\
\bottomrule
\end{tabularx}
\endgroup

\noindent\footnotesize Published or model-card nominal parameter counts are shown. The detailed model-inventory CSV separately preserves registry parameter estimates and local checkpoint tensor counts for provenance; manuscript parameter counts use the published or model-card values.
\normalsize

\newpage
\begin{landscape}
\subsection*{Supplementary Table 29. Manuscript-facing implementation settings}

\begingroup
\fontsize{6.8}{7.6}\selectfont
\setlength{\tabcolsep}{3pt}
\renewcommand{\arraystretch}{1.02}
\noindent\begin{tabularx}{0.98\linewidth}{p{0.24\linewidth}Y}
\toprule
Setting & Manuscript-facing value \\
\midrule
Blocked cross-validation unit & Ordered event blocks and participant-run or story/run grouping where matchable data permit it; manuscript-level inference uses participant-aware summaries where available. \\
Fold count and purge & Manuscript-facing targeted and closure recomputation uses three outer blocked folds and a one-sample purge where configured. \\
Ridge alpha grid & Ridge alpha is selected inside training data from 1.0, 10.0, and 100.0. \\
Lag/time-window grid & ECoG/iEEG windows: 0--100, 100--300, 300--600, 600--1000, and 0--1000 ms; MEG windows: 0--100, 100--250, 250--500, and 500--800 ms; no test-set lag or layer selection. \\
PCA dimensionality & Projection and PCA choices are train-only; manuscript-facing closure uses at most 32 components when dimensionality reduction is needed. \\
Hidden-state token pooling & Word-level hidden states use token-to-word maps with the final subtoken as the primary pooling rule. Mean-subtoken arrays are retained for audit or fallback, while the manuscript-facing predictive closure uses embedding, middle, and final layer groups with final-subtoken pooling. Fixed stores also record layer-indexed, early, and late groups where available. Feature extraction uses the configured 1024-token event context with automatic context reduction when needed; each checkpoint's native context window is retained in the inventory CSV. \\
Checkpoint IDs and revisions & \path{distilgpt2@fed02aafb76b46aa5da200966f9e42262e758023}; \path{gpt2@10c66461e4c109db5a2196bff4bb59be30396ed8}; \path{gpt2-medium@7b40acf43d912d857c663046ac178ee3fd7ff2d0}; \path{EleutherAI/pythia-160m@b8368ff94f3bcf3088de5e9912251fc0208ae524}; \path{EleutherAI/pythia-410m@9879c9b5f8bea9051dcb0e68dff21493d67e9d4f}; \path{Qwen/Qwen2.5-0.5B-Instruct@7ae557604adf67be50417f59c2c2f167def9a775}; \path{Qwen/Qwen2.5-1.5B-Instruct@989aa7980e4cf806f80c7fef2b1adb7bc71aa306}; \path{Qwen/Qwen2.5-7B-Instruct@a09a35458c702b33eeacc393d103063234e8bc28}; \path{Qwen/Qwen3-1.7B@70d244cc86ccca08cf5af4e1e306ecf908b1ad5e}; \path{Qwen/Qwen3-4B-Instruct-2507@cdbee75f17c01a7cc42f958dc650907174af0554}. \\
Subword-to-word aggregation & Word surprisal is the sum of subword surprisal values for each word using token-to-word maps when model-derived columns are available. \\
Score aggregation & Held-out scores are summarized at matched participant-run or target-profile units and then by dataset/model/layer/metric/candidate quantity. \\
Proxy policy & Proxy-derived quantities are source-boundary diagnostics and are excluded from model-specific computational correspondence support. \\
Synthetic implant strength & Strength is the coefficient \(\beta\) in \(y=\beta x_{\mathrm{implant}}+\epsilon\) for affected synthetic units on a unit-variance latent scale; the 80\% threshold is implementation-scale. \\
Aggregation path & Source inventories retain participant-run coverage. The participant-run predictive audit retains 26 Brain Treebank subject-run units from 10 participants, 44 MEG-MASC subject-run units from 11 participants, and 8 Podcast ECoG subject-run units from 8 participants after complete model-control matching. The integrated table is retained as a coverage-summary table. \\
\bottomrule
\end{tabularx}
\endgroup
\end{landscape}

\newpage
\begin{landscape}
\subsection*{Supplementary Table 30. Dataset-specific neural targets}

\begingroup
\fontsize{5.0}{5.6}\selectfont
\setlength{\tabcolsep}{1.0pt}
\renewcommand{\arraystretch}{0.78}
\noindent\begin{tabularx}{0.99\linewidth}{p{0.09\linewidth}p{0.15\linewidth}p{0.14\linewidth}p{0.12\linewidth}p{0.19\linewidth}p{0.11\linewidth}Y}
\toprule
Dataset & Neural measure and unit count & Exact target used & Window/profile & Selection rule & Final unit & Averaged before main contrast \\
\midrule
Brain Treebank & Intracranial ECoG electrode time series; source inventory maximum brain-unit count is 248. & Resolved HDF5 neural time-series keys plus sidecar electrode labels; all eligible electrodes are summarized, with no model-based target selection. & 0--1000 ms predictive summary; profile checks where required unit ordering is available. & Eligibility requires brain time series, word events, time table, model features, controls, and reliability metadata. & Participant-run predictive unit retained for 26 subject-run units from 10 participants. & Electrodes, target windows, and local rows are averaged within subject-run before predictive summaries, then runs are averaged within participant for cluster bootstrap. \\
Podcast ECoG & Intracranial ECoG/iEEG high-gamma derivative responses; source inventory maximum brain-unit count is 235. & Precomputed high-gamma electrode derivative time series aligned to podcast word events; all eligible electrodes are summarized, with no model-based target selection. & 0--1000 ms predictive summary; subject-to-subject brain-as-model profiles are auxiliary. & Eligibility requires derivatives, word events, model features, controls, and reliability metadata. & Participant-run predictive unit retained for 8 subject-run units from 8 participants. & Electrodes, target windows, and local rows are averaged within subject-run before predictive summaries. \\
MEG-MASC & KIT MEG sensor-space responses summarized as sensor-group targets; source inventory maximum brain-unit count is 257. & Sensor-space MEG channel time series from raw acquisition files summarized into sensor groups; no source-localized source-space target is used. & 100--250 ms predictive summary in the manuscript-facing audit; response-profile summaries use the matched target-profile grid. & Eligibility requires MEG target grid, word events, model-feature grid, matched controls, and reliability metadata. & Participant-run predictive unit retained for 44 subject-run units from 11 participants. & Sensor-group targets, model-feature rows, and profile units are averaged before dataset-level summaries, then runs are averaged within participant for cluster bootstrap. \\
\bottomrule
\end{tabularx}
\endgroup
\end{landscape}

\newpage
\subsection*{Supplementary Table 31. Aggregation-path audit}

\begingroup
\fontsize{7.0}{7.6}\selectfont
\setlength{\tabcolsep}{2.0pt}
\renewcommand{\arraystretch}{0.95}
\noindent\begin{tabularx}{0.98\linewidth}{p{0.18\linewidth}p{0.21\linewidth}p{0.22\linewidth}Y}
\toprule
Stage & Table & Identifier state & Consequence for inference \\
\midrule
Source inventory & \path{detailed_dataset_unit_inventory.csv} & Retains original subject, run, stimulus, unit type, and brain-unit coverage. & Shows source datasets contain multiple participants, runs, stimuli, and brain units. \\
Predictive score export & \path{detailed_predictive_scores.csv} & Includes subject/session/run/stimulus columns for model and control score rows. & Identifier columns are retained; participant-run inference depends on the matched model-control delta join. \\
Participant-run predictive audit & \path{main_table_participant_run_predictive_inference.csv} & Brain Treebank retains 26 subject-run units from 10 participants; MEG-MASC retains 44 subject-run units from 11 participants; Podcast ECoG retains 8 subject-run units from 8 participants. & All three primary datasets support participant-aware predictive summaries; repeated runs are averaged within participant before participant-cluster bootstrap. \\
Integrated coverage summary & \path{detailed_claim_evidence_decision_table.csv} & Joins predictive, response-profile, feature-ablation, ceiling, matched-control, and replication summaries into 504 coverage rows, including 432 evaluable predictive summary rows. & The integrated table provides coverage summaries; participant-run inference is reported separately for the predictive contrast. \\
Interval/equivalence audit & \path{main_table_participant_run_equivalence_audit.csv} & Reports retained participant counts, subject-run counts, participant-cluster bootstrap bounds where estimable, and unavailable coverage cases. & No SESOI was prespecified, so no formal equivalence claim is made. \\
\bottomrule
\end{tabularx}
\endgroup

\newpage
\subsection*{Supplementary Table 32. Predictive-only enrichment and participant-consistency coverage}

\begingroup
\scriptsize
\setlength{\tabcolsep}{2pt}
\renewcommand{\arraystretch}{1.02}
\noindent\begin{tabularx}{0.98\linewidth}{p{0.14\linewidth}p{0.12\linewidth}p{0.11\linewidth}p{0.18\linewidth}p{0.20\linewidth}Y}
\toprule
Dataset & Predictive-only rows & Models with passed rows & Layers with passed rows & Candidate quantities with passed rows & Participant consistency among passed configurations \\
\midrule
Brain Treebank & 38/144 & 8 & Embedding, middle, final & Semantic transition; context update & Median 5\%; range 0--30\% positive participant means \\
Podcast ECoG & 29/144 & 8 & Embedding, middle, final & Semantic transition; context update & Median 12.5\%; range 12.5--25\% positive participant means \\
MEG-MASC & 0/144 & 0 & None & None & Not applicable; no predictive-only rows \\
\bottomrule
\end{tabularx}
\endgroup

\noindent The corresponding machine-readable tables are available through the OSF data-and-code repository cited in the main text.

\newpage
\section*{Supplementary Table Index}

\begingroup
\scriptsize
\setlength{\tabcolsep}{3pt}
\renewcommand{\arraystretch}{1.05}
\noindent\begin{tabularx}{0.90\linewidth}{p{0.16\linewidth}Y}
\toprule
Item & Contents \\
\midrule
SI Table 1 & Dataset inventory and eligibility summary. \\
SI Table 2 & Evidence-chain closure and feasibility flags. \\
SI Table 3 & Binary criterion definitions and interpretation. \\
SI Table 4 & Matched controls. \\
SI Table 5 & Non-support and boundary labels. \\
SI Table 6 & Fixed source-package counts. \\
SI Table 7 & Integrated final decision and binary-criterion summary. \\
SI Table 8 & Claim-scope boundaries and audit rationale. \\
SI Table 9 & Model-side feature-ablation diagnostics by dataset. \\
SI Table 10 & Brain-positive control from brain-brain reliability. \\
SI Table 11 & Brain-as-model profile positive control. \\
SI Table 12 & Brain-profile unit-order implementation control. \\
SI Table 13 & Synthetic implanted-signal positive control and stochastic calibration. \\
\bottomrule
\end{tabularx}
\endgroup

\medskip
\begingroup
\scriptsize
\setlength{\tabcolsep}{3pt}
\renewcommand{\arraystretch}{1.05}
\noindent\begin{tabularx}{0.90\linewidth}{p{0.16\linewidth}Y}
\toprule
Item & Contents \\
\midrule
SI Table 14 & Independent low-level neural, standalone acoustic-envelope, full-length Podcast acoustic-envelope, Brain Treebank trial-to-audio mapping, and Brain Treebank official-release audit criteria. \\
SI Table 15 & Brain Treebank acoustic-envelope alternatives and claim scope. \\
SI Table 16 & Conventional-looking positives below matched-control criteria. \\
SI Table 17 & Dataset-level uncertainty summaries. \\
SI Table 18 & Interval-bound audit distinguishing coverage boundaries from equivalence tests. \\
SI Table 19 & Modality, region, and time-window predictive audit. \\
SI Table 20 & Feature-ablation uncertainty by dataset. \\
SI Table 21 & Criterion-threshold sensitivity. \\
SI Table 22 & Control-family ablation. \\
SI Table 23 & Brain-as-model oracle checks. \\
SI Table 24 & Leave-one-dataset-out final-decision sensitivity. \\
SI Table 25 & Leave-one-subject-run-out robustness. \\
SI Table 26 & Model inventory and coverage boundaries. \\
SI Table 27 & Detailed CSV file index. \\
SI Table 28 & Scope and limitation FAQ. \\
SI Table 29 & Manuscript-facing implementation settings for blocked cross-validation, ridge, lag/time-window handling, PCA, token pooling, subword aggregation, score aggregation, and proxy provenance. \\
SI Table 30 & Dataset-specific neural target, window, selection, and final-unit summary. \\
SI Table 31 & Aggregation-path audit from source inventory to decision table. \\
SI Table 32 & Positive-information participant summaries, predictive-only enrichment, and participant-consistency coverage. \\
\bottomrule
\end{tabularx}
\endgroup

\newpage
\section*{Detailed CSV Directory}

Detailed row-level tables are available through the OSF data-and-code repository cited in the main text. The arXiv package includes lightweight figure source-data exports as ancillary files under:

\begin{verbatim}
anc/figure_source_data/
\end{verbatim}

The positive-information, enrichment, and model-inventory tables used in Figure 1, Figure 2D, and Supplementary Table 32 are provided as:

\begin{itemize}
\item \texttt{main\_table\_positive\_information\_subject\_run.csv}
\item \texttt{main\_table\_positive\_information\_participant\_values.csv}
\item \texttt{main\_table\_positive\_information\_participant\_summary.csv}
\item \texttt{main\_table\_predictive\_only\_enrichment\_by\_dataset\_model\_layer.csv}
\item \texttt{main\_table\_predictive\_only\_enrichment\_overview.csv}
\item \texttt{main\_table\_predictive\_only\_participant\_consistency.csv}
\item \texttt{main\_table\_model\_inventory.csv}
\end{itemize}

\newpage
\section*{Claim Scope}

The supplement preserves the same scope as the main manuscript. Positive controls establish component-level sensitivity. Claims about language-model response-profile organization or candidate-computation correspondence require the calibrated contrasts applied to the tested model rows. Mean participant-run predictive model--control contrasts were nonpositive across the three primary datasets, although 67 of 432 evaluable predictive summary rows met the predictive-only criterion. Integrated mechanism-specific or reliability-bounded support was not established by the available contrasts. The integrated interpretation required jointly indexed measurements across all contrasts, which were not available in the current dataset coverage. Learning Brain records remain validation-only in this manuscript, with longitudinal plasticity and brain--LLM co-plasticity outside the current claim scope.

\end{document}